\documentclass[conference]{IEEEtran}
\IEEEoverridecommandlockouts
\usepackage{cite}
\usepackage{amsmath,amssymb,amsfonts}
\usepackage{algorithmic}
\usepackage{graphicx}
\usepackage{textcomp}
\usepackage{xcolor}
\usepackage{float}
\usepackage{epstopdf}
\usepackage{epsfig}
\usepackage{subfigure}
\usepackage{cleveref}
\usepackage[linesnumbered,ruled,vlined]{algorithm2e}

\newtheorem{definition}{Definition}

\def\BibTeX{{\rm B\kern-.05em{\sc i\kern-.025em b}\kern-.08em
    T\kern-.1667em\lower.7ex\hbox{E}\kern-.125emX}}
\begin{document}

\title{Residual-Quantile Adjustment for Adaptive Training of Physics-informed Neural Network\\
\thanks{978-1-6654-8045-1/22/\$31.00 ©2022 IEEE}
}

\author{\IEEEauthorblockN{Jiayue Han}
\IEEEauthorblockA{\textit{School of Data Science} \\
\textit{City University of Hong Kong}\\
Hong Kong, China \\
jyhan5-c@my.cityu.edu.hk}
\and
\IEEEauthorblockN{Zhiqiang Cai$^*$}
\IEEEauthorblockA{\textit{School of Data Science} \\
\textit{City University of Hong Kong}\\
Hong Kong, China \\
zqcai3-c@my.cityu.edu.hk}
\and
\IEEEauthorblockN{Zhiyou Wu}
\IEEEauthorblockA{\textit{Department of Mathematics} \\
\textit{City University of Hong Kong}\\
Hong Kong, China \\
zhiyouwu2-c@my.cityu.edu.hk}
\and
\IEEEauthorblockN{Xiang Zhou}
\IEEEauthorblockA{\textit{ School of Data Science and Department of Mathematics  } \\
\textit{City University of Hong Kong}\\
Hong Kong, China \\
xiang.zhou@cityu.edu.hk}
}

\maketitle

\begin{abstract}
Adaptive training methods for Physics-informed neural network (PINN) require dedicated constructions of the distribution of weights assigned at each training sample. 
To efficiently seek such an optimal weight distribution is not a simple task and most existing methods choose the adaptive weights based on approximating the full distribution or the maximum of residuals. In this paper, we show that the bottleneck in the adaptive choice of samples for training efficiency is the behavior of the tail distribution of the numerical residual. Thus, we propose the Residual-Quantile Adjustment (RQA) method for a better weight choice for each training sample. After initially 
setting the weights proportional to the $p$-th power of the residual, our RQA method 
reassign all weights above $q$-quantile ($90\%$ for example) to the median value, so that the weight follows a quantile-adjusted distribution derived from the residuals.   
This iterative reweighting technique,  on the other hand, is also very easy to implement. Experiment results show that the proposed method can outperform several adaptive methods on various partial differential equation (PDE) problems.
\end{abstract}


% \fi
\IEEEpeerreviewmaketitle

%%%%%%%%%%%%%%%%%%%%%%%%%%%%%%%%%%%%%%%%%%%%%%%%%%%%%%%%%%%%%%%%%%%%%%%%%%%%%%%%%%%%%
\section{Introduction}
Solving partial differential equations (PDEs) has been a long-standing inevitable problem when handling real-world applications of physics and engineering. Though some of the PDEs can be solved analytically, most of them rely on numerical methods to deal with. Traditional methods such as the finite element method and finite difference method cannot tackle large-scale and high dimensional problems thus using deep neural network (DNN) in PDE computation has become a popular research topic these years \cite{sirignano2018dgm,Raissi2019,zang2020weak,yu2018deep}.

A recent line of work on using physics-informed neural networks (PINN) \cite{Raissi2019} to solve PDEs has received widespread attention. This work approximates the solution of PDEs with a neural network which takes in $x$ and $t$ as input and outputs the solution $u(x,t)$. The squared $L_2$-norm (usually in Lebesque measure) of the residual for the PDEs is taken as the loss function, and the initial conditions and boundary conditions are added to the loss as penalty terms. 

Albeit successful in many fields, PINN may get stuck into a trivial solution as a local minimum \cite{krishnapriyan2021characterizing,daw2022rethinking}. The residuals for the PDEs equal to 0 does not mean finding the solution to the PDEs, because there are lots of trivial solutions that satisfy the condition that the residuals for the PDEs are 0. Adding boundary conditions and initial conditions can help PINN to escape from the trivial solutions. However, if the PDEs are really irregular, then PINN may face propagation failures characterized by highly imbalanced PDEs residual fields, where very high residuals are observed in very narrow regions of the domain \cite{daw2022rethinking}. 
% However, the computation time for this method is quite long. This is because the optimization process contains two part in this work. For the first part, Adam is applied to obtain a course solution and for the second part BFGS is used to obtain a better result. 

% To improve the efficiency of Neural-network based PDE solvers, various frameworks are proposed. In Weak Adversarial Network (WAN) \cite{zang2020weak}, Zhang et al. introduced the test function and obtain the weak formulation of PDEs. The weak solution is approximated by a primal neural network as in PINN, while an adversarial network is trained simultaneously to represent the test function. At each iteration, the adversarial net aims at maximizing the loss function and the primal net aims at minimizing it. As a result, adversarial net may help accelerate the convergence.

Adaptive techniques are an efficacious approach to avoid the failure of PINN. Identifying the high-residual points in the domain can be used to prevent the propagation failure of PINN.  For instance, Selectnet \cite{gu2021selectnet} is one of these works but relies on one extra neural network taking the role of weight at each training point to discern the importance of each point.  However, Selectnet needs to solve a min-max problem
for both the weighted network and the solution network. Meanwhile, in theory, the weight function in that min-max problem is only supported  on the locations with the maximum value of residuals. Some careful training practice to balance both networks is critical for the success of efficient  adaptive training.

% Selectnet \cite{gu2021selectnet} is one work that improve the convergence speed of network-based least square method for PDEs. As in WAN, it also introduce a new auxiliary network, Selectnet, to help accelerate the convergence by maximizing the loss function at each iteration. Selectnet takes in $x$ and $t$ and outputs a weight for this $(x,t)$ pair which stand for the "importance" of the data. The weighted sum of the residual over the whole domain is then used as the loss function. 
% In our work, we show that this extra network in Selectnet is actually unnecessary. With a weight designed according to residual, we can obtain an adaptive scheme for network-based PDE solver. This attempt, however, may introduce singularity on the area with large residual. So we apply an quantile regression adjustment: for the weights larger than a certain quantile of all the weights in the dataset, we reset them to the median of all weights. Our method is compared to Selectnet over three numerical examples and our method outperforms Selectnet on all of them. 

The simplest and heuristic choice for the weights 
is to follow the value of the residuals. In the traditional grid-based adaptive PDEs method, a very classic and successful refinement strategy of adjusting mesh size is exactly based on this choice.  On the other hand, the min-max adversarial adaptive training
in Selectnet \cite{gu2021selectnet} suggests a theoretically singular distribution
supported only on maximum values. However, for network-based PDEs solvers like PINN, directly assigning the weights proportional to the residuals or on the maximum values are both far away from the satisfactory performance.

In this work, we first choose the weight of each data point according to iterative reweighted least square (IRLS), which is setting weights to the normalized $(p-2)$-th $(p\geq 2)$ power of the residuals. A higher $p$ means we pay more attention to samples with larger residuals. Tuning parameter $p$ is problem-dependent and challenging since a minuscule $p$ may nullify the adaptivity while a too large $p$ may deteriorate the performance. 
 
To mitigate the amplifying effect of the singularity of a large $p$ value, we propose the Residual-Quantile Adjustment algorithm (RQA) to adjust the weight. Specifically, it selects the data points with residuals larger than a certain quantile of all residuals and adjusts their weights to another moderate-size value (for example the median, how to choose this value is discussed in Section \ref{sec:e}). This innovative design is inspired by the observation that when  singularities occur  in the domain, they are usually very concentrated.
Since the residuals take  a very high value near these singular locations,
the distribution of the residuals  typically has a long tail, which 
prevents the efficiency decrease in the mean square error. 
If the sample weight distribution would literally follow  this residual distribution, 
most of the weights will be concentrated on  very few data points near the  singularity 
while failing to provide the necessary attention to other bulk training samples. 
This effect is even worse if the dimension of
the PDE is higher. Our adjustment based on suppressing the weight of data points with residuals larger than a certain threshold can effectively avoid such an “overweighing” issue
and achieve adaptivity robustly and
efficiently.

Though simple, this RQA idea is proved to be useful. Extensive experiments show that after using quantile adjustment, the performance of using $L_p$-norm induced weight improves remarkably.
We state the main contribution for this work as follows:
\begin{itemize}
    \item RQA provides an iterative reweighting technique for network-related PDE solvers without introducing new optimization problems.
    \item The quantile adjustment strategy can be easily applied to any residual-based training algorithms.
    \item The insight of adaptivity is also reveal in this work: Adaptivity not only means to give larger weights to the worse learned data, but also involves balancing their weights with that of the well learned ones.
    \item By RQA, we demonstrate that  with the same   least square loss as   the residual's (squared)  mean,  the residual {\it quantile} is a good indicator in concerns of
    training efficiency. 
\end{itemize}

%*****************
% In this work, we present that the maximizer for the ``selection network" in the Selectnet is just the $L_p$-nrom of the residual. It means that we do not need to train the min-max problem and it will reduce the computation cost for the training of Selectnet. Actually, different $p$ will lead to different performance of the training of PINN. We compare the performance of different $p$ and investigate the adaptive method for PINN.

% We show that using $L_p$-norm of the residual of the PDE as the selection network will lead to the singularity of the weights of the training point. The singularity means that it may put very large weight into the high-residual points and almost very small weight into the ``insignificant" points. So the histogram of the weights have ``long tail". Then the training of PINN will focus on the high-residual points and overlook the ``insignificant" points. 

% We propose a Residual-Quantile Adjustment algorithm (RQA) to prevent ``long tail" phenomenon. The numerical experiments for a variety of PDE problems show that our proposed method have the better performance.
%******************

The paper is organized as follows. Section \ref{sec:rw} is a brief introduction to the related work for network-based PDE solvers. Problem formulation is mentioned in Section \ref{sec:pf}. Section \ref{sec:main} states our main method RQA including the details for $L_p$-norm induced weight and the quantile adjustment. In Section \ref{sec:e}, we compare RQA with other adaptive methods and discuss the choice of parameters through numerical experiments. Section \ref{sec:con} ends the paper with a brief conclusion and future works.

%%%%%%%%%%%%%%%%%%%%%%%%%%%%%%%%%%%%%%%%%%%%%%%%%%%%%%%%%%%%%%%%%%%%%%%%%%%%%%%%%%%%%
\section{Related Work}\label{sec:rw}
\textbf{Physics-informed neural network (PINN).}
Using deep neural networks (DNNs) to solve partial differential equations (PDEs) has been developed rapidly. As far as we know, Deep Galerkin Method (DGM) \cite{sirignano2018dgm} is the first method that uses DNNs to solve the PDEs. Then in 2019, Maziar et al. \cite{Raissi2019} proposed the most well-known PINN, which approximates the solution of PDEs with a neural network. The loss function is designed based on the residual, while initial conditions and boundary conditions are added as penalty terms. However, PINN may face failure when the PDEs are really irregular \cite{krishnapriyan2021characterizing}. Thus, a series of extensions of PINN has been proposed to boost the performance of PINN from various aspects. 

% Thus, WAN \cite{zang2020weak} considers the weak form of the PDE solution and use the theory of the adversarial network to solve the PDEs. 

% Many network-based PDE solvers may suffer from several issues in practice. 

\textbf{Different loss functions.}
Apostolos et al. \cite{psaros2022meta} proposed a meta-learning method to propose  an alternative loss function for PINN. Gradient-enhanced PINN \cite{yu2022gradient} combines the gradient information of the PDE residual into the loss function. Instead, Zeng et al. \cite{zeng2022adaptive} used the adaptive loss function introduced in \cite{barron2019general} to improve the performance of PINN.

\textbf{Time-dependent problems.} For time-dependent problems, dividing the time segments into small time steps and training the neural networks from small time steps and then extending it into the long time steps can be helpful \cite{zhao2020solving,krishnapriyan2021characterizing,mattey2022novel,haitsiukevich2022improved,wang2022respecting}. If the problem is in the large domain, it is suggested to decompose the spatio-temporal domain to accelerate the training of PINN \cite{meng2020ppinn,shukla2021parallel,jagtap2021extended}. 

\textbf{Weight-based adaptive methods.} The weights in each term of the loss function are constants in PINN, and these weights will affect the convergence of PINN. Tuning the weights in each term of the losses and balancing the losses will improve  the performance of PINN \cite{wang2021understanding,wang2022and,xiang2022self}. Besides considering the weights in each term of the losses, extensive works focus on  the  weights for a given set of training points and most of them  assign more weights to training points with high residual values \cite{lu2021physics,mcclenny2020self,gu2021selectnet,li2022revisiting}. 

\textbf{Sampling-based adaptive methods.} The continuous strategy 
in parallel to assigning weights to a fixed number of training points is to propose a distribution density function to sample training points. Such a distribution function can also be called as a weight function. Selectnet \cite{gu2021selectnet} trains the neural network to learn this function as the weight. In solving the Fokker-Planck equation, Tang et al.\cite{tang2021pinns} directly use the
  solution of the Fokker-Planck equation as the weight function and learn a generative model to sample training points.  
   Also, some other sampling-based methods use Markov chain Monte Carlo  to sample the residual distribution as the weight distribution \cite{nabian2021efficient,gao2021active, wu2022comprehensive,hanna2022residual}.  Other sampling-based methods include\cite{wu2022comprehensive,daw2022rethinking,jagtap2021extended,peng2022rang}.
   However, these methods need extra non-trivial computational overhead in order to obtain the weight function or generate samples and few of them paid enough attention to the choice of the distribution function itself.  
 
\textbf{Augmented Lagrangian methods.} Besides the methods above, a series of improvements have been done to PINN based on the augmented Lagrangian method. In these works, instead of adding the initial condition and boundary conditions as penalty terms, they are considered as the constraints of an optimization problem whose objective function is the $l_2$ loss of the residual on the training set. R. Mojgani et al. proposed the Lagrangian PINN \cite{lagrangepinn} framework which split the original residual into two parts and uses different networks to calculate them. S. Basir designed PECANNs \cite{pecann} and a method based on dual problem \cite{dual} to make better use of the Lagrangian multiplier methods. These methods are especially efficient in dealing with failure modes of PINN for example the convection equation with a large coefficient for the $\frac{{\rm d}u}{{\rm d}x}$ term.

%%%%%%%%%%%%%%%%%%%%%%%%%%%%%%%%%%%%%%%%%%%%%%%%%%%%%%%%%%%%%%%%%%%%%%%%%%%%%%%%%%%%%
\begin{figure*}[htbp]
\centering
\includegraphics[width=0.8\textwidth]{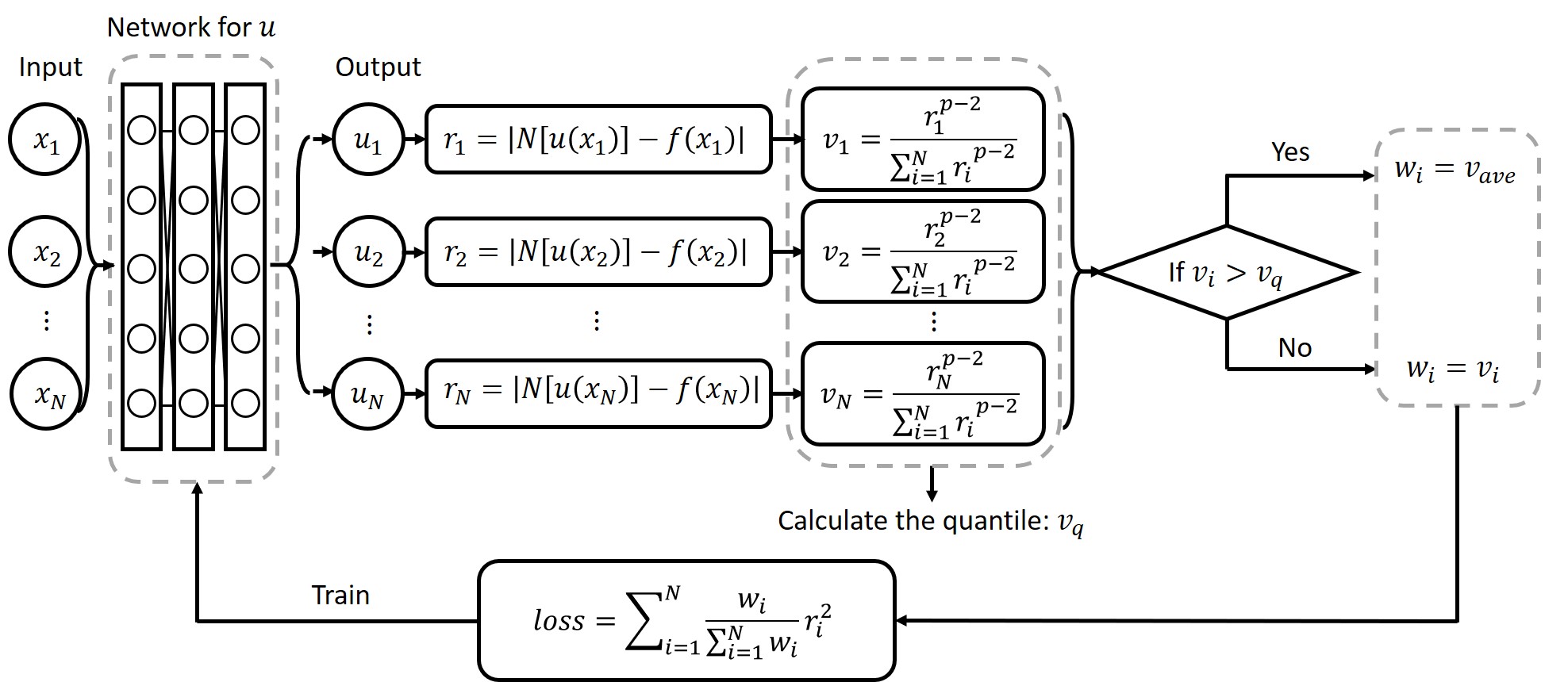}
 \caption{Framework of our main method RQA.}
  \label{fig:main_method}
\end{figure*}

\section{Problem Formulation}
\label{sec:pf}
Consider the PDE of the form:
\begin{align}
\mathcal{N}[u]=f(x,t), x \in \Omega, t \in [0,T],
\end{align}
where $u(\cdot,\cdot):\Omega\times [0,T]\to \mathbb{R}$ is the PDE solution, $\mathcal{N}[\cdot]$ is a linear or nonlinear differential operator, and $\Omega$ is a domain in $\mathbb{R}^{d}$.

Let $r(x,t)$ be the residual of the PDE at $(x,t)$,
\begin{align}r(x,t)=\left|\mathcal{N}[u(x,t)]-f(x,t)\right|.\end{align}

As in Physics-informed neural networks (PINN) \cite{Raissi2019}, we parameterize the solution $u(x,t)$ by a neural network $u(x,t;\theta)$, where $\theta$ denotes the parameters. The solution for PDE is then obtained by optimizing parameters in the following problem.
\begin{align}
\min_\theta \int_{[0,T]}\int_\Omega r^2_\theta(x,t){\rm d}x{\rm d} t\approx \frac{1}{N}\sum_{i=1}^N r^2_\theta(x_i,t_i), 
\end{align}
where $r_\theta(x,t)=|\mathcal{N}[u(x,t;\theta)]-f(x,t)|$ and $\{x_i,t_i\}_{i=1}^N$ is the dataset sampled uniformly in the domain $\Omega\times [0,T]$. 

Furthermore, if there exists boundary condition $\mathcal{B}[u]=g(x,t)$ on $\partial \Omega\times[0,T]$ or initial condition $\mathcal{I}[u]=h(x)$ on $\Omega$, these conditions would be added in the form of penalty.
\begin{align}\min_{\theta}&\frac{1}{N} \sum_{i=1}^N r^2_\theta(x_i,t_i),\notag\\
&+\frac{\lambda_B}{N_B} \sum_{j=1}^{N_B} \left|\mathcal{B}[u(x_j^{(B)},t_j^{(B)})]-g(x_j,t_j)\right|^2,\notag\\
&+\frac{\lambda_I}{N_I}  \sum_{k=1}^{N_I} \left|\mathcal{I}[u(x_k^{(I)},0)]-h(x_k)\right|^2,
\end{align}
where  $\{x_j^{(B)},t_j^{(B)}\}_{j=1}^{N_B}$ is uniformly sampled on the boundary $\partial \Omega\times[0,T]$ and $\{x_k^{(I)}\}_{k=1}^{N_I}$ is uniformly sampled in $\Omega$, $\lambda_B$ and $\lambda_I$ are two hyperparameters.

Although the mesh-free PINN approach can deal with high-dimensional PDEs problems, the convergence of this method is slow. To deal with it, Gu et al. proposed an adaptive accelerating method known as the Selectnet. Selectnet introduces an auxiliary network $\phi_r(x,t;\theta_r)$ to select the weights for each sample point $(x,t)$. 

The optimization problem to be solved is the min-max problem:
\begin{equation}\label{eqn:selectnet}
    \min_\theta\max_{\theta_r} \int_{[0,T]}\int_\Omega \phi_r(x,t;\theta_r) r_\theta^2(x,t) {\rm d}x {\rm d}t,
\end{equation} 
subject to the normalization condition
\begin{align}\frac{1}{T|\Omega|}\int_{[0,T]}\int_\Omega \phi_r(x,t;\theta_r){\rm d}x{\rm d}t=1.\end{align}
As in most machine-learning methods, the constraints are added as the penalty terms. So the objective function becomes: 
\begin{align}\label{eqn:selectnet2}
    \min_\theta\max_{\theta_r} &\int_{[0,T]}\int_\Omega \phi_r(x,t;\theta_r) r_\theta^2(x,t) {\rm d}x{\rm d}t\notag\\ &-\frac{1}{\epsilon}\left(\frac{1}{T|\Omega|}\int_{[0,T]}\int_\Omega \phi_r(x,t;\theta_r){\rm d}x{\rm d}t-1\right)^2.
\end{align}
This adaptive approach achieves the goal of self-paced learning. Meanwhile, since neural networks tend to capture low-frequency information, Selectnet can avoid the singularity in other adaptive methods, such as the binary weight method as shown in the paper.

\section{Main Methods}
\label{sec:main}
The main framework of RQA can be seen in Figure \ref{fig:main_method}. In Section \ref{subsec:adaptic_p}, we propose the maximizer of the selection network in Selectnet and derive the $L_p$-norm induced weights adaptive method for PINN. Section \ref{subsec:quantile_a} shows the quantile adjustment technique to balance the $L_p$-norm induced weights.

\subsection{Adaptive through $L_p$-norm}
\label{subsec:adaptic_p}
From equation \eqref{eqn:selectnet}, we can see that in order to solve the PDE problems, Selectnet needs to conduct a min-max optimization. This kind of optimization is usually treated very carefully in the network-based method, and all the parameters should be chosen properly. Also, training an additional network is costly. We show that we can actually avoid the use of Selectnet by the deduction below.

Consider the objective function
\begin{align}
\min_\theta\max_{\theta_r} \int \phi_r (x,t;\theta_r)^{\frac{1}{q}} r_\theta(x,t) {\rm d}x{\rm d}t,
\end{align}
subject to $\int \phi_r(x,t;\theta_r) {\rm d}x{\rm d}t=1$. 
    
We have
\begin{equation}
\begin{split}
&\int \phi_r(x,t;\theta_r)^{\frac{1}{q}} r_\theta(x,t) {\rm d}x{\rm d}t,\\
&=\langle \phi_r(x,t;\theta_r)^{\frac{1}{q}},r_\theta(x,t)\rangle,\\
&\leq \|\phi_r(x,t;\theta_r)^{\frac{1}{q}}\|_q\|r_\theta(x,t)\|_p,\\
& = \left(\int \phi_r(x,t;\theta_r) {\rm d}x\right)^{\frac{1}{q}}\|r_\theta(x,t)\|_p,\\
& = \|r_\theta(x,t)\|_p,\label{equ:p-norm}
\end{split}
\end{equation}
where $\frac{1}{p}+\frac{1}{q}=1$, $p,q\geq 1$.

As a result, we may get rid of the extra network that approximates weight in Selectnet. 
    
Notice that in \eqref{equ:p-norm}, if $p=2$, then this is exactly the traditional least square method. In the following contents, we suppose that $p\geq2$.

Taking $\{x_i,t_i\}_{i=1}^N$ as the dataset, minimizing $\|r_\theta(x,t)\|_p$ is then approximated by minimizing 
\begin{align}\min_\theta \left(\sum_{i=1}^{N}|r_\theta(x,t)|^p\right)^{\frac{1}{p}},\end{align} which is equivalent to
\begin{align}\min_\theta \sum_{i=1}^{N}|r_\theta(x,t)|^p,\end{align} and the latter is easier to compute. 

Let $\hat{\theta}$ be some parameter close to $\theta$ (for example, in iterative methods, $\hat{\theta}$ can be chosen as the value of $\theta$ from the previous iteration), the loss function above can be approximated by
\begin{align}\min_\theta \sum_{i=1}^{N}|r_{\hat{\theta}}(x_i,t_i)|^{p-2}|r_\theta(x,t)|^2.\end{align} This approach obviates the potential computation difficulty if $p$ is not an integer.

Finally, we conduct normalization and state the objective function of the $p$-norm based adaptive method as follows:
\begin{align}\min_\theta \sum_{i=1}^N \frac{|r_{\hat{\theta}}(x_i,t_i)|^{p-2}}{\sum_{i=1}^N|r_{\hat{\theta}}(x_i,t_i)|^{p-2}}|r_\theta(x_i,t_i)|^2.\end{align}

As above, if the PDE problem contains boundary conditions and initial conditions, we add them as the penalty term. We first denote the residuals as below:
\begin{align}
    r^{(B)}(x,t)&=\left|\mathcal{B}[u(x,t)]-g(x,t)\right|,\\
    r^{(I)}(x,t)&=\left|\mathcal{I}[u(x,0)]-h(x)\right|.
\end{align}
And the loss function is 
\begin{equation}
\begin{split}
&\min_\theta \sum_{i=1}^N \frac{|r_{\hat{\theta}}(x_i,t_i)|^{p-2}}{\sum_{i=1}^N|r_{\hat{\theta}}(x_i,t_i)|^{p-2}}|r_\theta(x_i,t_i)|^2\\&
+ \lambda_B \sum_{i=1}^{N_B} \frac{|r^{(B)}_{\hat{\theta}}(x_i,t_i)|^{p-2}}{\sum_{i=1}^{N_B}|r^{(B)}_{\hat{\theta}}(x_i,t_i)|^{p-2}}|r^{(B)}_\theta(x_i,t_i)|^2\\
&+\lambda_I \sum_{i=1}^{N_I} \frac{|r^{(I)}_{\hat{\theta}}(x_i,t_i)|^{p-2}}{\sum^{N_I}_{i=1}|r^{(I)}_{\hat{\theta}}(x_i,t_i)|^{p-2}}|r^{(I)}_\theta(x_i,t_i)|^2.
\end{split}
\end{equation}
All the notations are the same as above. Note that the $p$ value for each term can be different. For simplicity, we do not distinguish them here.

\subsection{Quantile adjustment}
\label{subsec:quantile_a}
The $L_p$-norm based method proposed above, however, will directly introduce ``singularity", especially when $p$ is very large. This is because during the training process, the residual for the flat area will become relatively smaller, and the residual for the singular area will become relatively larger. A large $p$ value will intensify this difference thus resulting in a larger difference in weight.  

\begin{figure}[htbp]
\centering
\includegraphics[scale=0.5]{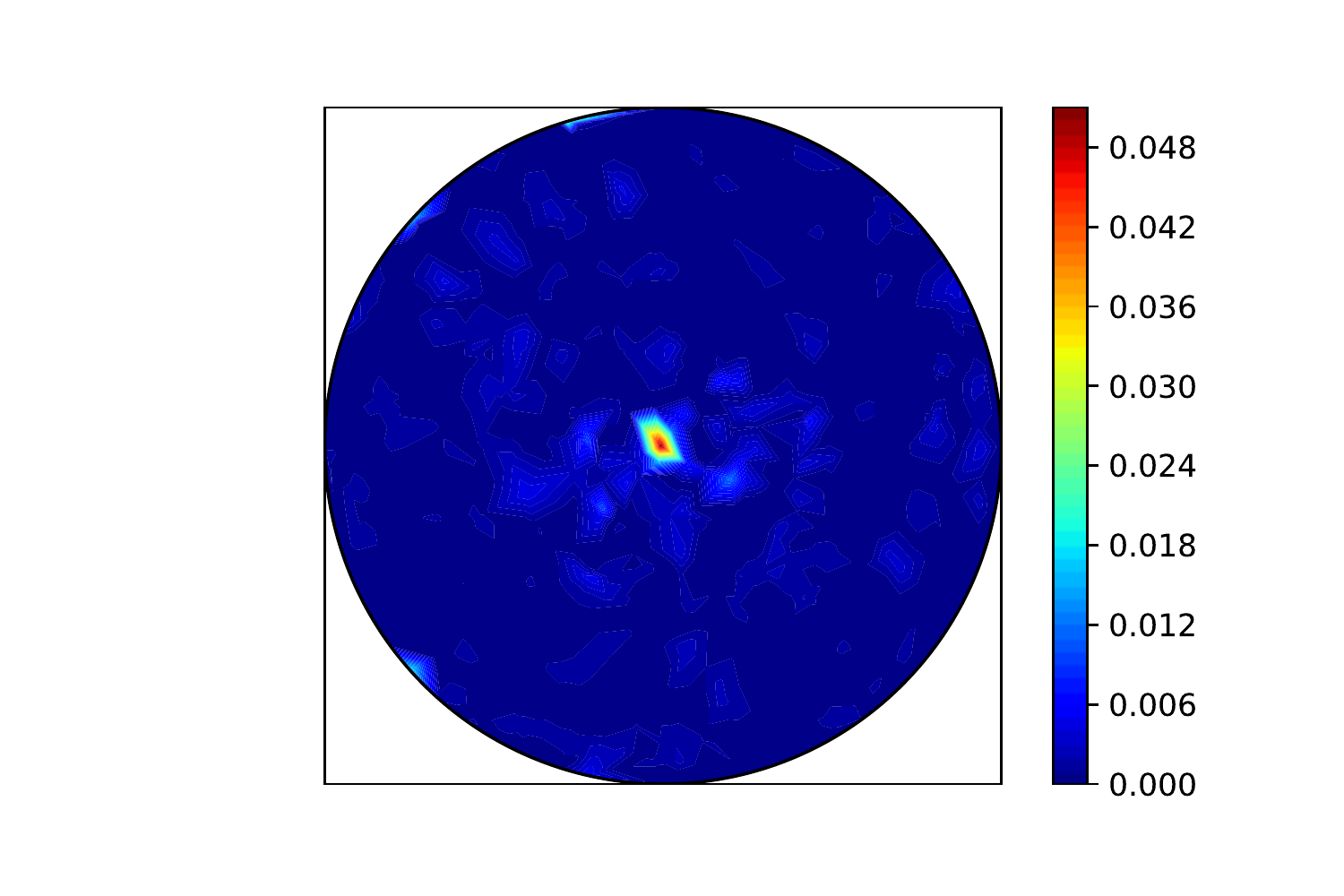}
\caption{Weights induced by the $L_p$-norm after 1000 iterations, $p=3$.}
\label{fig:singularity}
\end{figure}

For example, Figure \ref{fig:singularity} shows the $L_p$-norm induced weight for PDE problem (\ref{eq:prob1}) after 1000 iterations. It can be observed that the weight of the center area is much more larger than the rest parts, which makes the data points in the middle more ``important". 

Specifically, the $(x,t)$ pairs with larger residuals in the dataset may take up all the weight, and other $(x,t)$ pairs can hardly contribute to the loss function.  As a result, the maximum error on the domain may be small, but the average error will be large. As a consequence, the adaptive method with $L_p$-norm induced weight may converge slowly.
% Notice that the advantage of using neural network to approximate the weight is that network can avoid singularity. The $p$-norm based method proposed above, however, will directly introduce singularity, especially when $p$  is large. Which means that the $(x,t)$ pair with largest residual in the dataset may take up all the weight and other $(x,t)$ pairs can hardly contribute to the loss function. As a result, the maximum error on the domain may be small but the average error will be large.

% \begin{figure}[H]
% \centering

% \setcounter{subfigure} {0} {\includegraphics[scale=0.28]{1000_N_points_iter1.eps}}
% \setcounter{subfigure} {0} {\includegraphics[scale=0.28]{1000_N_points_iter1080.eps}}
% \caption{\it{Left:} Initial weights induced by the $p$-norm based method. \it{Right:} Weights induced by the $p$-norm after 1000 iterations.}
% \label{fig:singularity}
% \end{figure}

% consider the mixture distribution of the residual-based distribution and the uniform distribution. 

\begin{figure*}[htbp]
\centering
\subfigure{
\begin{minipage}[t]{0.4\linewidth}
\centering
    \includegraphics[width=\textwidth]{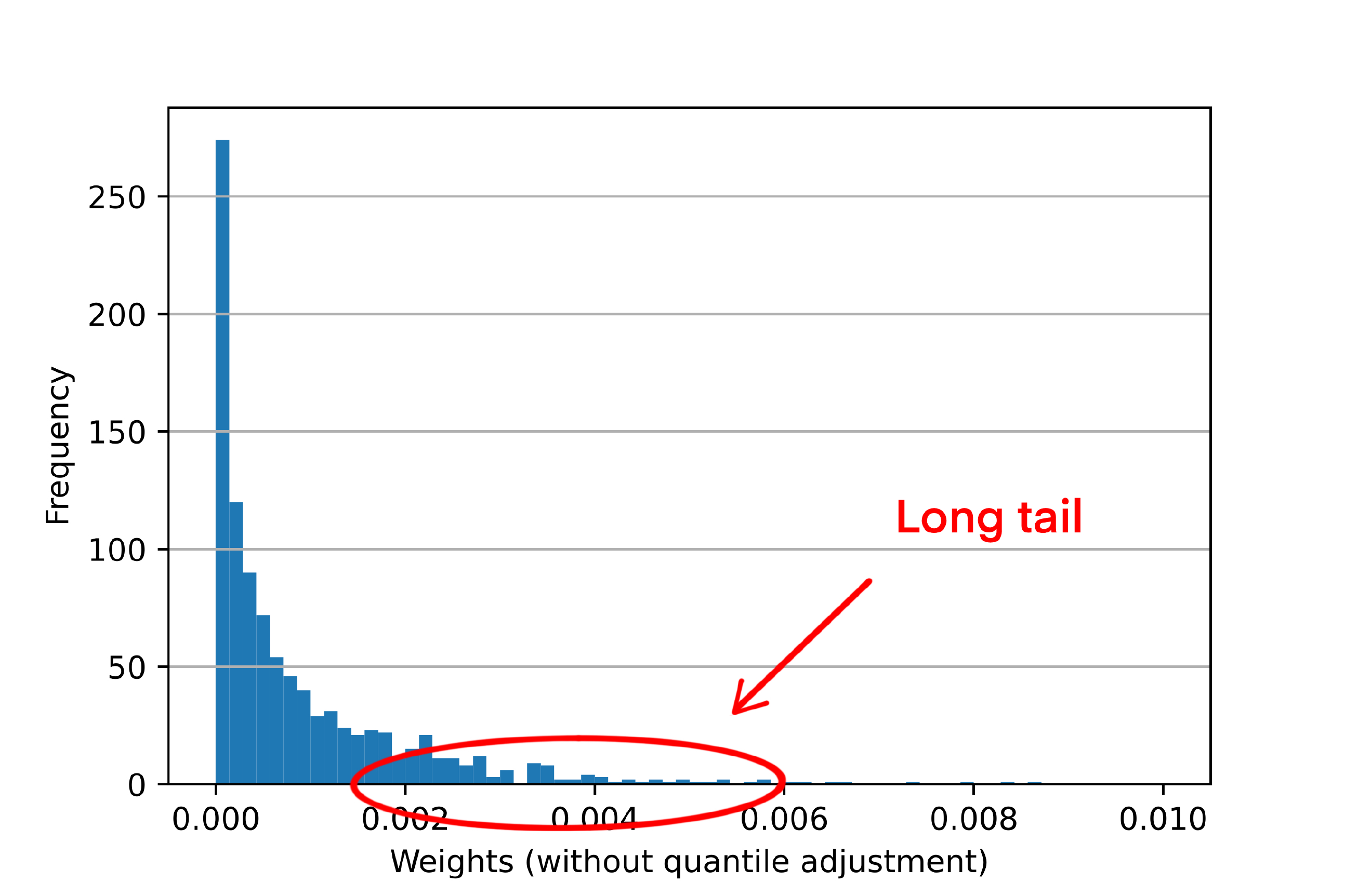}
    \caption{Histogram of the weights induced by the $L_p$-norm based method.}
    \label{fig:hist_w_n_q}
    \end{minipage}
}
\subfigure{
\begin{minipage}[t]{0.4\linewidth}
\centering
    \includegraphics[width=\textwidth]{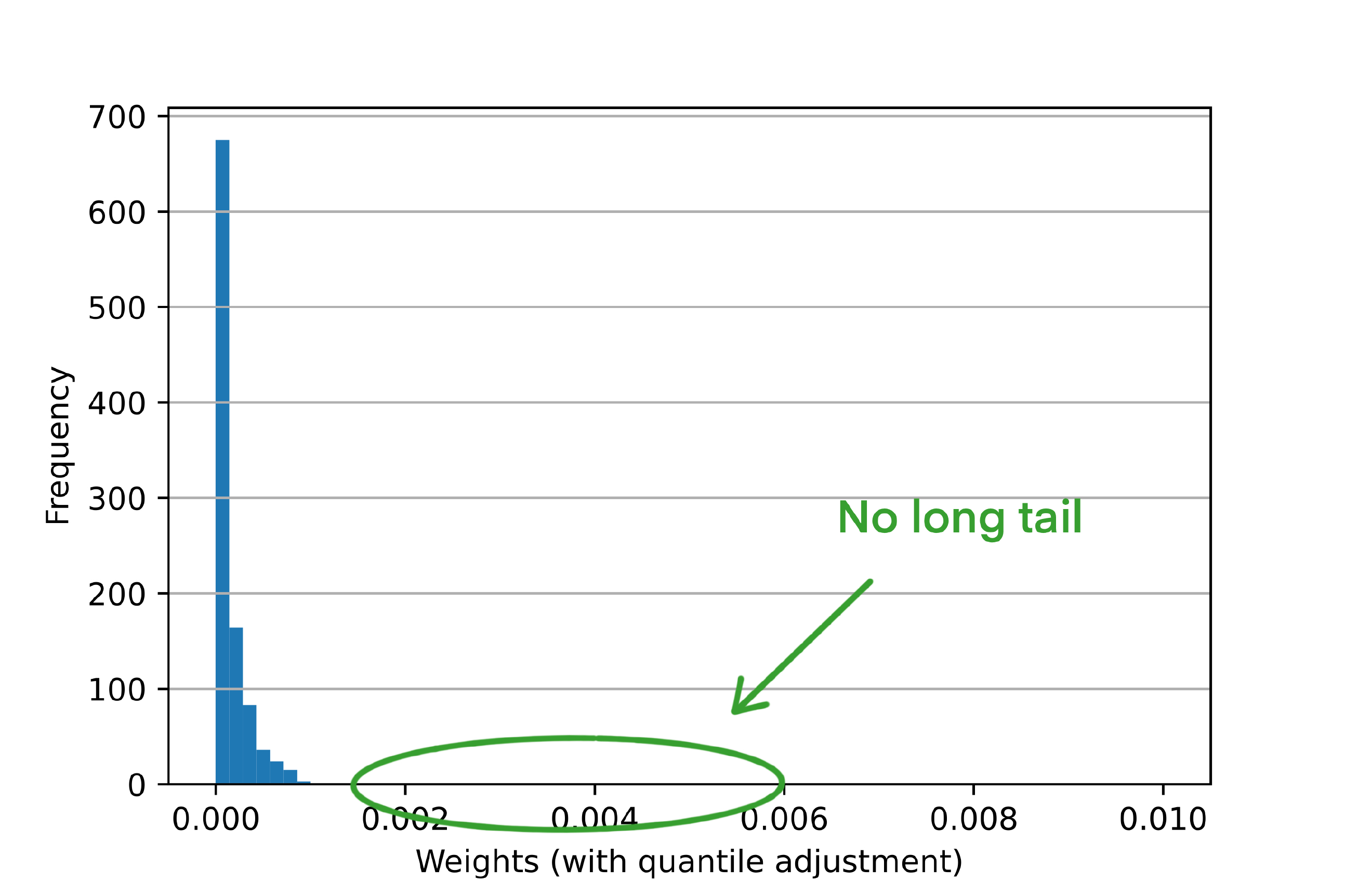}
    \caption{Histogram of the weights induced by the $ L_p$-norm based method with quantile adjustment.}
    \label{fig:hist_w_q}
    \end{minipage}
}
\subfigure{
\begin{minipage}[t]{0.4\linewidth}
\centering
    \includegraphics[width=\textwidth]{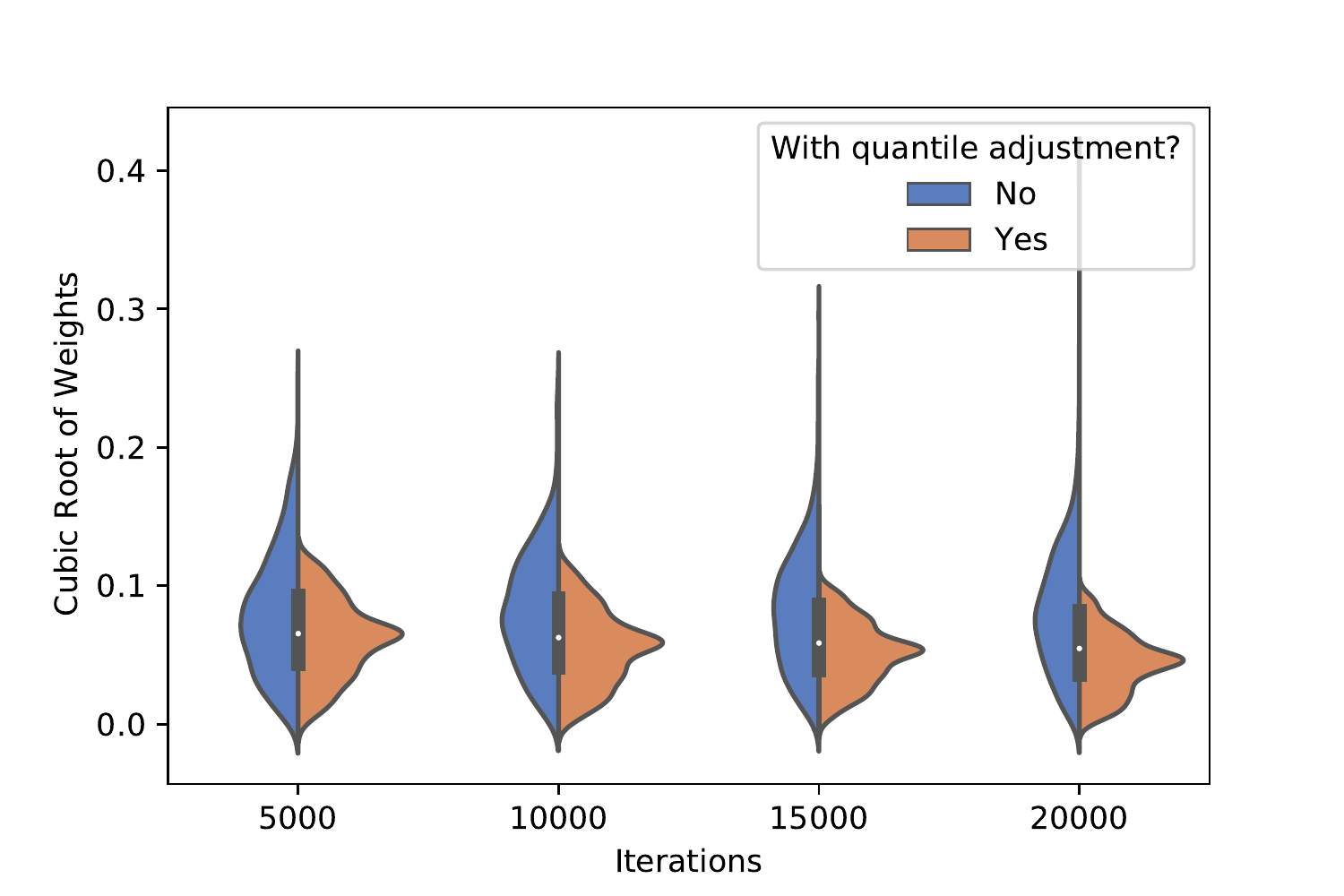}
    \caption{Violin plot of the weights for different iterations. The blue one represents the weights without quantile adjustment and the orange one represents the weights with quantile adjustment.}
    \label{fig:violin}
    \end{minipage}
}
\end{figure*}

To avoid the case that the loss function overlooks the data points with low weights (small residual), a straightforward strategy is to reallocate the weights of the singular area to the rest.
Note that, ``singularity" can be visualized as the ``long tail" phenomenon in the histogram of the weights as shown in Figure \ref{fig:hist_w_n_q}. Therefore if we can eliminate the ``long tail" phenomenon, then ``singularity" can be solved. It means that the loss function can focus on the data points with large residuals without ignoring the data points with small residuals.

%----------------The three figures---------------------
% \begin{figure}[H]
% \centering
% \includegraphics[width=0.4\textwidth]{Without_quantile_histogram_20000_iteration.pdf}
% \caption{Histogram of the weights induced by the $p$-norm based method}
%  \label{fig:hist_w_n_q}
% \end{figure}

% \begin{figure}[h]
% \centering
% \includegraphics[width=0.4\textwidth]{With_quantile_histogram_20000_iteration.pdf}
%  \caption{Histogram of the weights induced by the $p$-norm based method with quantile adjustment.}
%  \label{fig:hist_w_q}
% \end{figure}

% \begin{figure}[H]
% \centering
% \includegraphics[width=0.45\textwidth]{Cubicroot_weights.eps}
% \caption{Violin plot of the weights for different iterations. The blue one represents the weights without quantile adjustment and the orange one represents the weights with quantile adjustment.}
% \label{fig:violin}
% \end{figure}
%------------------------------------------------------

We now introduce quantile to deal with the ``long tail" in the histogram. 

% To describe the ``long tail" in the histogram, quantile can be the suitable quantity. 

\begin{definition}
    The $\xi$-th quantile for the cumulative distribution function $F$ is defined as $F^{-1}(\xi)=\inf\{x: F(x)\leq \xi\}$ for $\xi\in(0,1)$.
\end{definition}

% Note that different $\xi$ can measure the dividing line between the ``important" data points and ``unimportant" data points.

We first select a specific quantile of the weight.

For the weights smaller than the quantile, their weights are considered to be reasonable, so they remain the same; for the weights larger than the quantile, their weights are considered to be too large, so we adjust the weights to a proper number (for example, the median of all weights). The new weights are then normalized afterward. We name this method the "Residual-Quantile adjustment" (RQA).

Figure \ref{fig:hist_w_q} shows the result of our RQA, we can observe that the ``long tail" has been cut down. With Residual-Quantile adjustment, all the data including those with notably large or small residuals can be taken care of during the training, as illustrated in Figure \ref{fig:violin}.  The numerical examples in the next section show that RQA performs better than directly using the $L_p$-norm induced weight and a more intricate method Selectnet \cite{gu2021selectnet}.

% combine the information of the uniform distribution into the $L_p$-norm residual-based weights

% \section{Numerical Details}
% \label{sec:nm}
% The numerical details are listed as below:
% \begin{itemize}
%     \item Optimization method: ADAM \cite{kingma2014adam}.
%     \item Total iteration: n=5000.
%     \item Training data: $N=500$, $N_B=500$.
%     \item Test data: $N_t=10000$.
%     \item Stepsize:
%     \begin{itemize}
%         \item For selection network: $\tau_s^{(k)}=10^{-4}$
%         \item For solution network: 
%         \begin{itemize}
%             \item Example 1 and Example 3:
%             $$\tau^{(k)}=10^{-2-3j/1000}, n^{(j)}<k\leq n^{(j+1)}, \forall j=0,...,1000$$
%             \item Example 2:
%             $$\tau^{(k)}=10^{-3-3j/1000}, n^{(j)}<k\leq n^{(j+1)}, \forall j=0,...,1000$$
%         \end{itemize}
%         where $0=n^{(0)}<\cdots<n^{(1000)}=n$ are equidistant segments of total iterations.
%     \end{itemize}
%     \item Network Structure: 
%     \begin{itemize}
%         \item For solution network, we use 3-layer network with activation function $\sigma(x)=\max (x^3,0)$. At each layer, we have 100 neurons.
%         \item For selection network, we use 3-layer ReLu network. At each layer, we have 20 neurons.
%     \end{itemize}
    
%     \item Hyperparameters: 
%     \begin{itemize}
%         \item $p=3$ for both data inside the domain and data on the boundary.
%         \item $\lambda_B=1$.
%         \item $\epsilon=0.001.$
%     \end{itemize}
    
%     \item Quatile: The 10-quantiles. 

%\end{itemize}

\section{Experiment}\label{sec:e}
The experiment section consists of three parts. In the first part, we compare our method with other adaptive methods for network-based PDE solvers. In the second part, we apply our method to high-dimensional PDEs. The selection of parameters in our algorithm is discussed in the last part.

\subsection{Compare with other methods}\label{sec:exp1}
In this experiment, we compare RQA with three other adaptive methods:
\begin{itemize}
    \item $L_p$-norm induced weight.
    \item Binary weighting.
    \item Selectnet.
\end{itemize}
The adaptive method with $L_p$-norm induced weight can be considered as our method without quantile adjustment. Binary weighting is an adaptive method that separates the dataset into 2 parts and assigns elements of the same dataset a certain constant as their weights. Specifically, a dataset with $N$ data is separated into $\eta N (0<\eta<1)$ data with larger residuals and $(1-\eta)N$ data with smaller residuals. Weight for the former is $w_L>1$ and weight for latter is $w_S<1$, these weights satisfy the constraint $w_L\eta+w_S(1-\eta)=1$. Here we choose $\eta=0.8, w_L/w_S=4$, which is the best pair of parameters as shown in Selectnet.

In terms of the solution network, we use 3-layer fully connected neural network (FCN) with activation function $\sigma(x)=\max (x^3,0)$ (100 neurons for each layer). This structure of networks is applied to all experiments. The step size $\tau^{(k)}$ at the $k$-th iteration is defined as follows:
  $$\tau^{(k)}=10^{-2-3j/1000}, n^{(j)}<k\leq n^{(j+1)}, \forall j=0,...,1000.$$   
where $0=n^{(0)}<\cdots<n^{(1000)}=n$ are $1000$ equidistant segments of total $n$ iterations. 

We need to mention that in many other network-based PDE solvers, a special network structure is used to make the solution satisfy the boundary condition automatically. Here, to show the capability of RQA, we just apply FCN and learn the boundary condition by adding penalty terms. Higher accuracy can be achieved if using a specifically designed network structure. 
 
For the Selectnet, we use the same network as in the original paper, a 3-layer ReLu network with 20 neurons for each layer. The step size is fixed as $\tau_s^{(k)}=10^{-4}$.

In our method, we set $p=3$ for all the data inside the domain, on the boundary, and in the initial condition. The quantile adjustment for Section \ref{sec:exp1} and Section \ref{sec:exp2} refers to setting the weight larger than $90\%$ quantile to $50\%$ quantile, and then we conduct normalization. A detailed discussion on these parameters can be found in Section \ref{sec:exp3}.

The data points allocation of the training set is 1000 points inside the domain, 1000 points on the boundary, and 50 points for $t=0$, for all problem settings.
 
Other parameters are set to be: $\lambda_B=1$, $\lambda_I=1$ and $\epsilon=0.001$.
At each iteration, we record the $L_2$-error and the absolute maximum loss of a test set with $N_t=10000$ points uniformly selected from the domain.
\begin{itemize}
    \item $L_2$-error
    \begin{align}\sqrt{\frac{\sum^{N_t}_{i=1}(u(x_i,t_i)-u(x_i,t_i;\theta))^2}{\sum^{N_t}_{i=1}u(x_i,t_i)^2 }}, i=1,...,N_t;\end{align}
    \item Absolute maximum error
    \begin{align}\frac{\max_i\left|u(x_i,t_i)-u(x_i,t_i;\theta)\right|}{\max_i\left|u(x_i,t_i)\right|}, i=1,...,N_t.\end{align}
\end{itemize}
where $u(x,t)$ is the true PDE solution at $(x,t)$, $u(x,t;\theta)$ is the solution approximated by neural network at $(x,t)$, $\{x_i,t_i\}_{i=1}^{N_t}$ is the test set.

\subsubsection{Linear parabolic equation}\label{sec:lpe}
\begin{figure}[H]
\centering
\includegraphics[width=0.4\textwidth]{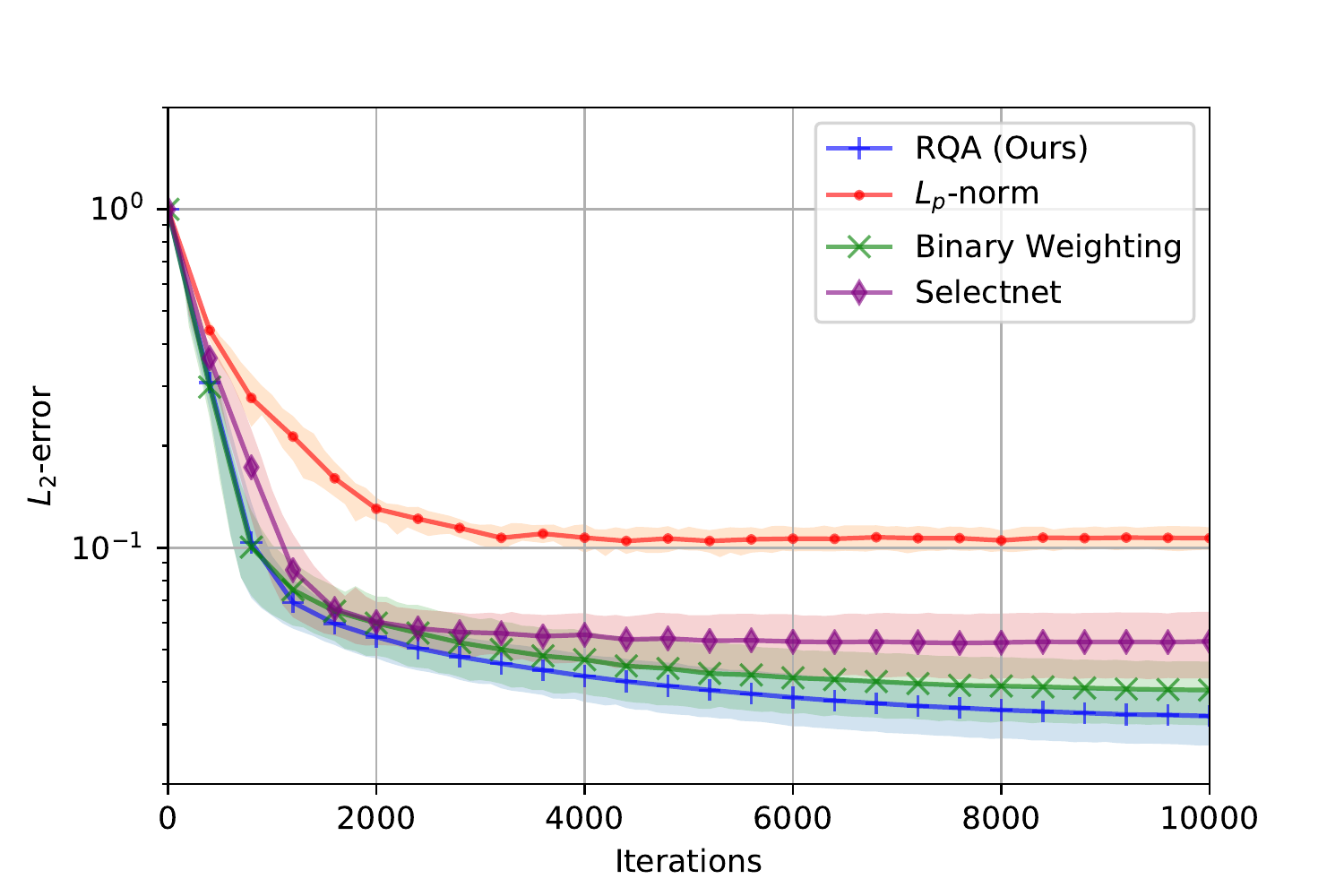}
 \caption{Comparison of $L_2$-error for various adaptive methods at each iteration for the 5-dim linear parabolic equation.}
\label{Fig:2-5-l2}
\end{figure}

\begin{figure}[H]
\centering
\includegraphics[width=0.4\textwidth]{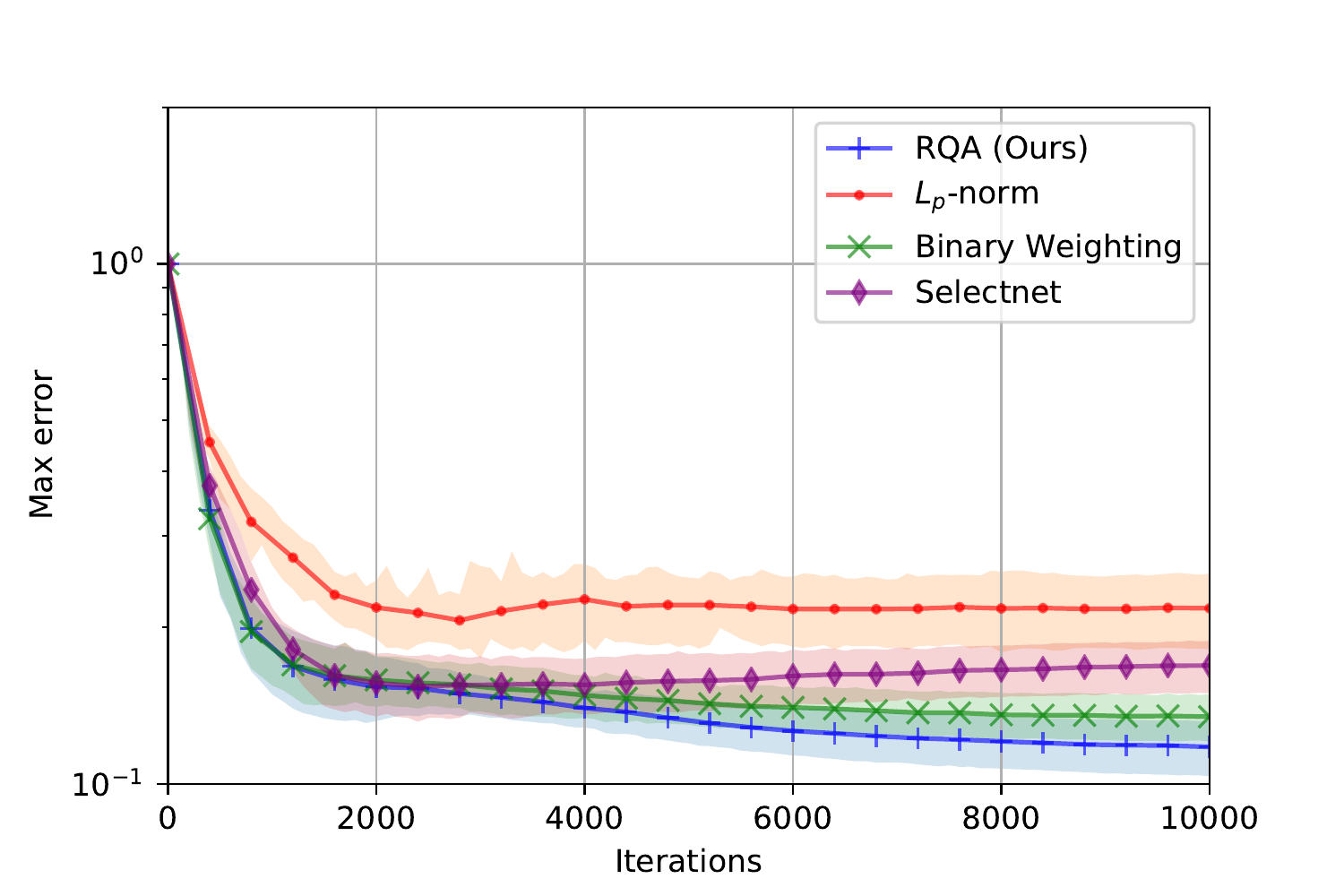}
 \caption{Comparison of absolute maximum error for various adaptive methods at each iteration for the 5-dim linear parabolic equation ("Max error" is short for "Absolute maximum error").}
\label{Fig:2-5-m}
\end{figure}

For the first problem, we consider the linear parabolic equation, the problem setting is stated below:
\begin{equation}
\label{eq:lpe}
\begin{split}
\mathcal{N}_1[u(x,t)]&=f(x, t), \quad \text { in } Q:=\Omega \times(0,1), \\
u(x, t) &=g(x,t), \quad \text { on } \partial \Omega \times(0,1), \\
u(x, 0) &=h(x), \quad \text { in } \Omega,
\end{split}
\end{equation}
with the exact solution
\begin{align}
u(x,t)=\exp(\|x\|_2\sqrt{1-t}),
\end{align}
and
\begin{align}
    \mathcal{N}_1[u(x,t)]&=\partial_{t} u(x, t)-\nabla_{x} \cdot\left(a(x) \nabla_{x} u(x, t)\right),
\end{align}

\begin{align}
f(x,t)=&-0.5u(x,t)\log{(u(x,t))}\notag\\
&-u(x,t)\log{(u(x,t))}\notag\\
&-(1+0.5(\log{(u(x,t))})^2)u(x,t)(1-t)\\
&+\sqrt{1-t}(d-1)/\|x\|_2),\notag\\
g(x,t)=&\exp(\sqrt{1-t}),\\
h(x)=&\exp{(\|x\|_2)},
\end{align}
where $a(x)=1+\frac{1}{2}\|x\|_2^2$ and $\Omega:=\{x:\|x\|_2<1\}$.
Let the problem dimension be 5, we run the four adaptive methods mentioned above (RQA, $L_p$-norm induced weight, binary weighting, Selectnet) on 10 different runs. At each run, the initials for the solution network, the training set, and the test set for each iteration are identical. The training process takes 10000 iterations. Means of $L_2$-error for all methods are shown in Figure \ref{Fig:2-5-l2} with lines and the shaded areas in the figure representing the area between $[mean-std,mean+std]$.
We also plot the absolute maximum loss for all methods in Figure \ref{Fig:2-5-m}, all legends are the same as in Figure \ref{Fig:2-5-l2}.

From the figure, we can see that RQA (represented in the red line) has the smallest $L_2$-error and the absolute maximum error. The binary weighting method (represented in green line) and Selectnet (represented in purple line) have the second and third best results. Using $L_p$-norm induced weight only, on the other hand, is the worse. It not only converges the slowest but also obtains the highest error. However, the only difference between RQA and $L_p$-norm induced weight is the use of quantile adjustment. This result shows that the quantile adjustment does help to avoid singularity and guarantee the performance of RQA.

%=================================================

\subsubsection{Allen-Cahn equation}\label{sec:ac}
For the second problem, we consider the famous Allen-Cahn equation
\begin{equation}
\begin{split}\label{eq:ac}
\mathcal{N}_2[u(x,t)]&=f(x, t), \quad \text { in } Q:=\Omega \times(0,1), \\
u(x, t) &=g(x,t), \quad \text { on } \partial \Omega \times(0,1), \\
u(x, 0) &=h(x), \quad \text { in } \Omega,
\end{split}
\end{equation}
and the exact solution is 
\begin{align}
u(x,t)=e^{-t}\sin(\frac{\pi}{2}|1-\|x\|_2|^{2.5}),
\end{align}
and
\begin{align}
\mathcal{N}_2[u(x,t)]=&\partial_{t} u(x, t)-\Delta_{x} u(x, t)\notag\\
&-u(x, t)+u^{3}(x, t),\\
g(x,t)=&0,\\
h(x)=&\sin(\frac{\pi}{2}|1-\|x\|_2|^{2.5}),\\
l(x)=&\cos(0.5\pi|1-\|x\|_2|^{2.5}),\\
k(x)=&-\frac{5}{4}\pi(d-1)\|x\|_2l(x)|1-\|x\|_2|^{1.5}\notag\\
&-\frac{25}{16}\pi^2h(x)(1-\|x\|_2)^3\\
&+\frac{15}{8}\pi l(x)|1-\|x\|_2|^{0.5},\notag\\
f(x,t)=&-e^{-t}(h(x)+k(x))\notag\\&-u(x,t)+u(x,t)^3.
\end{align}
where $\Omega:=\{x:\|x\|_2<1\}$.

\begin{figure}[H]
\centering
\includegraphics[width=0.4\textwidth]{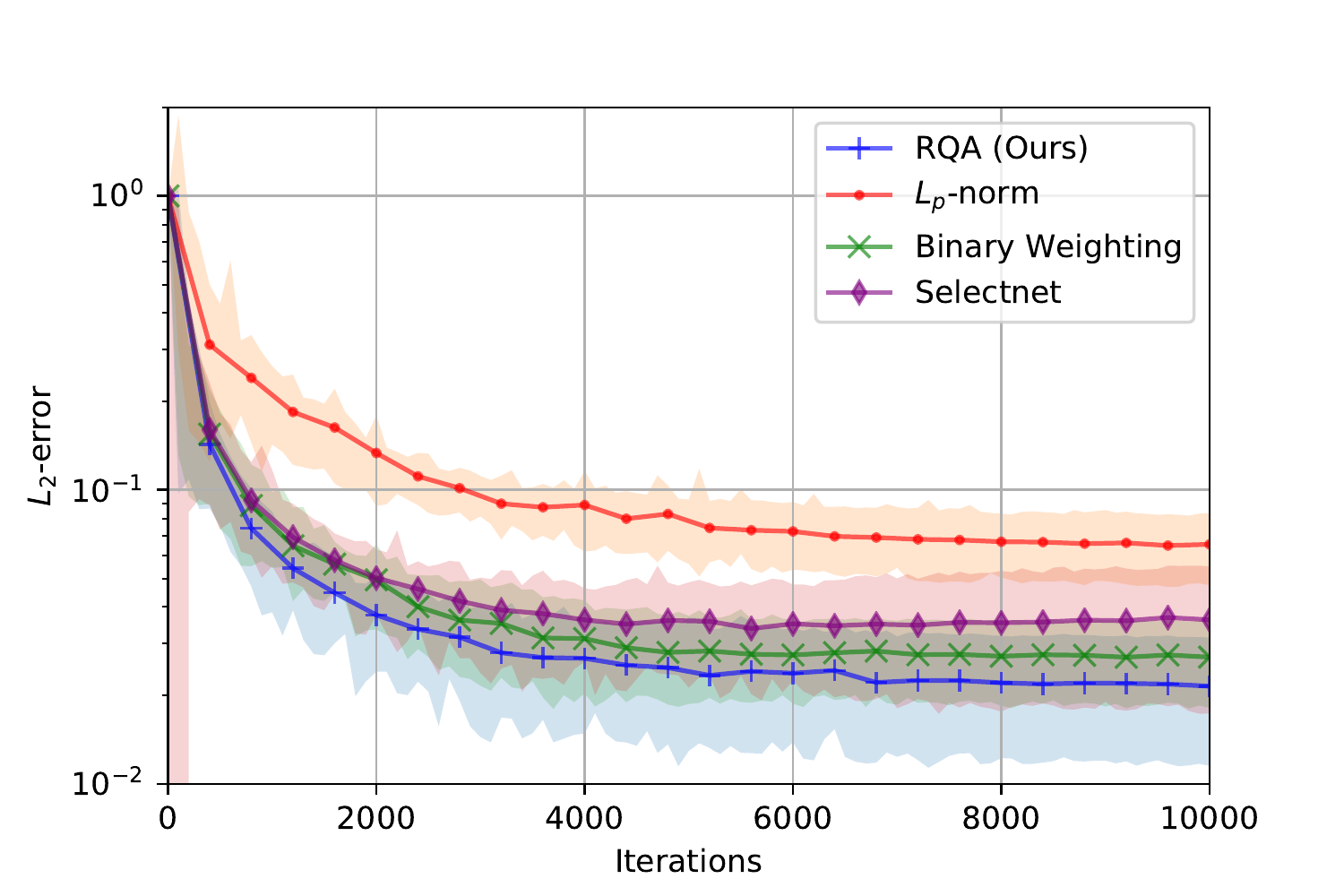}
\caption{Comparison of $L_2$-error for various adaptive methods at each iteration for 5 dim Allen-Cahn equation.}
\label{Fig:3-5-l2}
\end{figure}

\begin{figure}[H]
\centering
\includegraphics[width=0.4\textwidth]{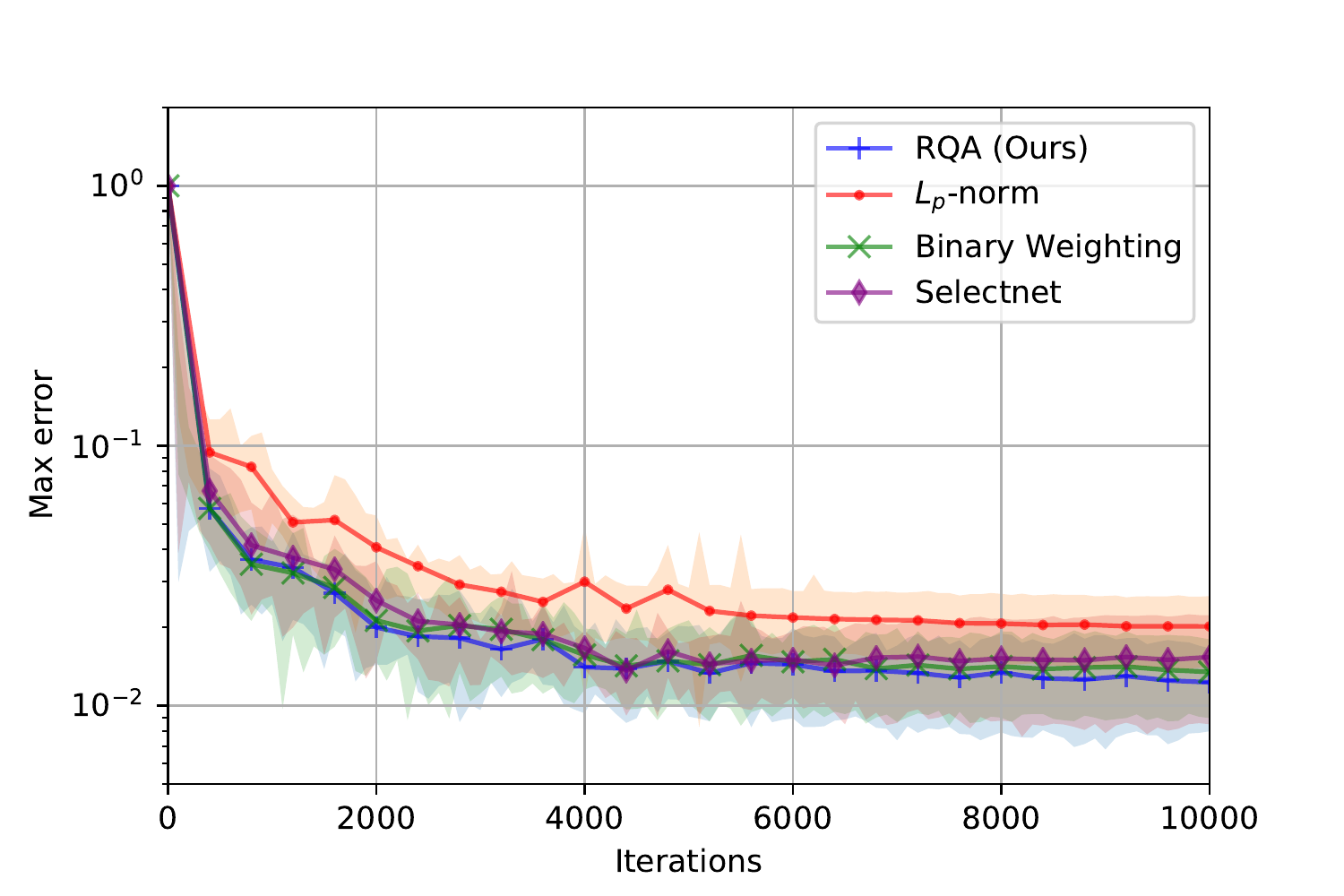}
 \caption{Comparison of absolute maximum error for various adaptive methods at each iteration for 5 dim Allen-Cahn equation.}
\label{Fig:3-5-m}
\end{figure}

The dimension setting for this problem is also 5. Figure \ref{Fig:3-5-l2} and Figure \ref{Fig:3-5-m} show the mean and standard deviation of $L_2$-error and absolute maximum error over 10 repeated experiments. Our method still performs the best, followed by the binary weighting method and Selectnet. Using $L_p$ induced norm only is still the worst. This result is similar to the linear parabolic equation, which shows the generalization of our method.
\begin{figure*}[htbp]
\centering
\subfigure{
\begin{minipage}[t]{0.4\linewidth}
\centering
    \includegraphics[width=\textwidth]{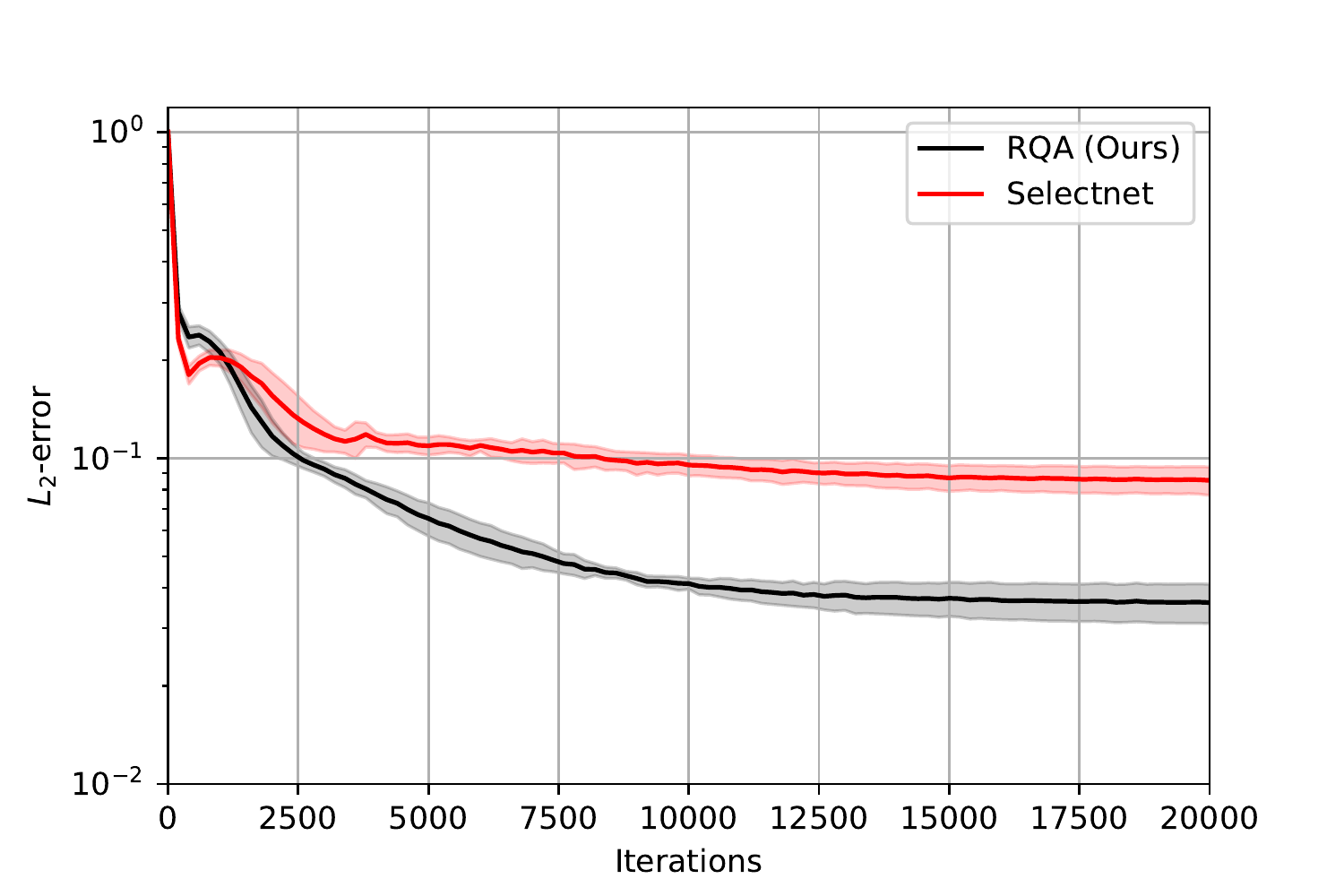}
    \caption{Comparison of $L_2$-error between RQA and Selectnet for 20 dim linear parabolic equation.}
    \label{Fig:2-20-l2}
    \end{minipage}
}
\subfigure{
\begin{minipage}[t]{0.4\linewidth}
\centering
    \includegraphics[width=\textwidth]{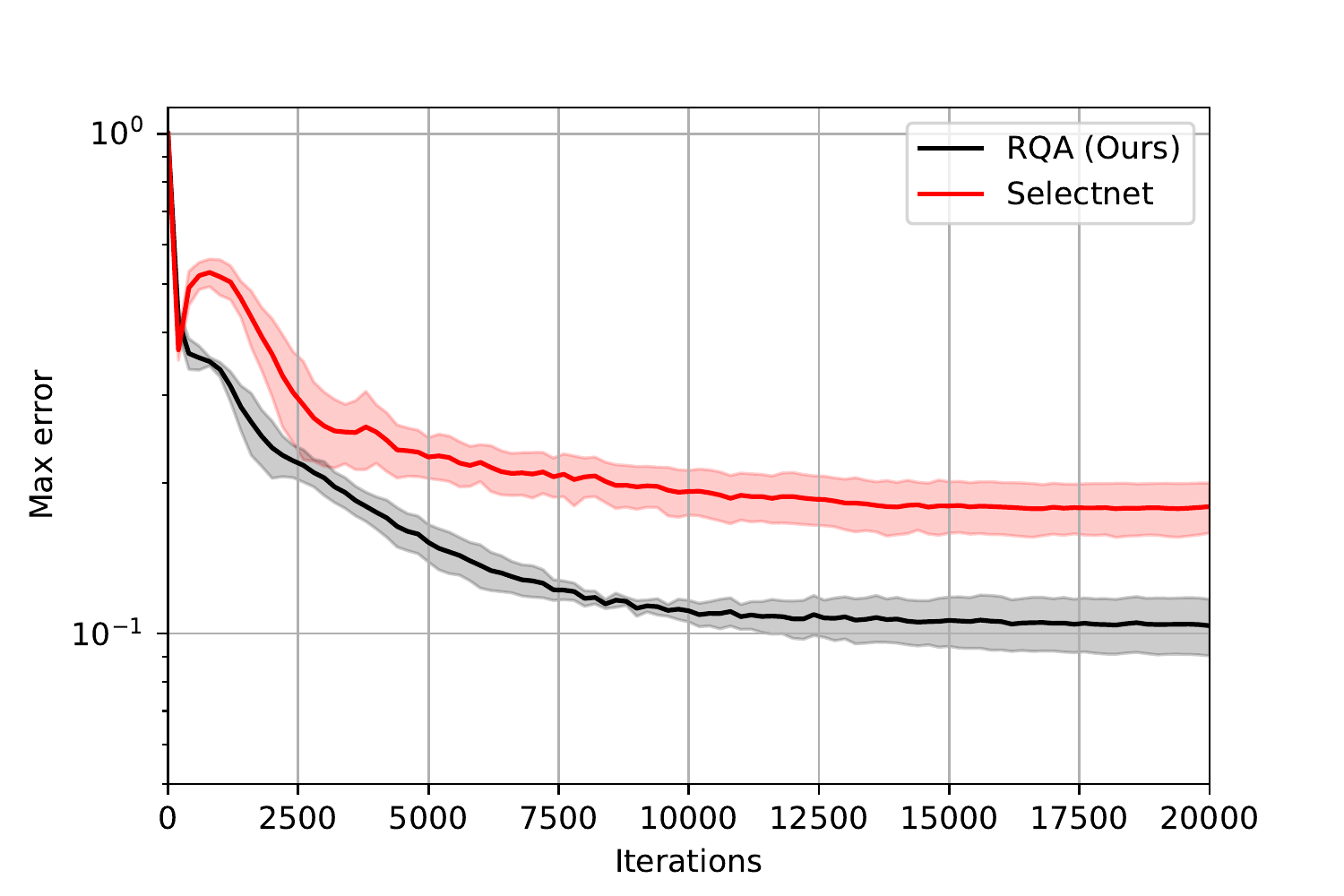}
    \caption{Comparison of absolute maximum error between RQA and Selectnet for 20 dim linear parabolic equation.}
    \label{Fig:2-20-m}
    \end{minipage}
}

\subfigure{
\begin{minipage}[t]{0.4\linewidth}
\centering
    \includegraphics[width=\textwidth]{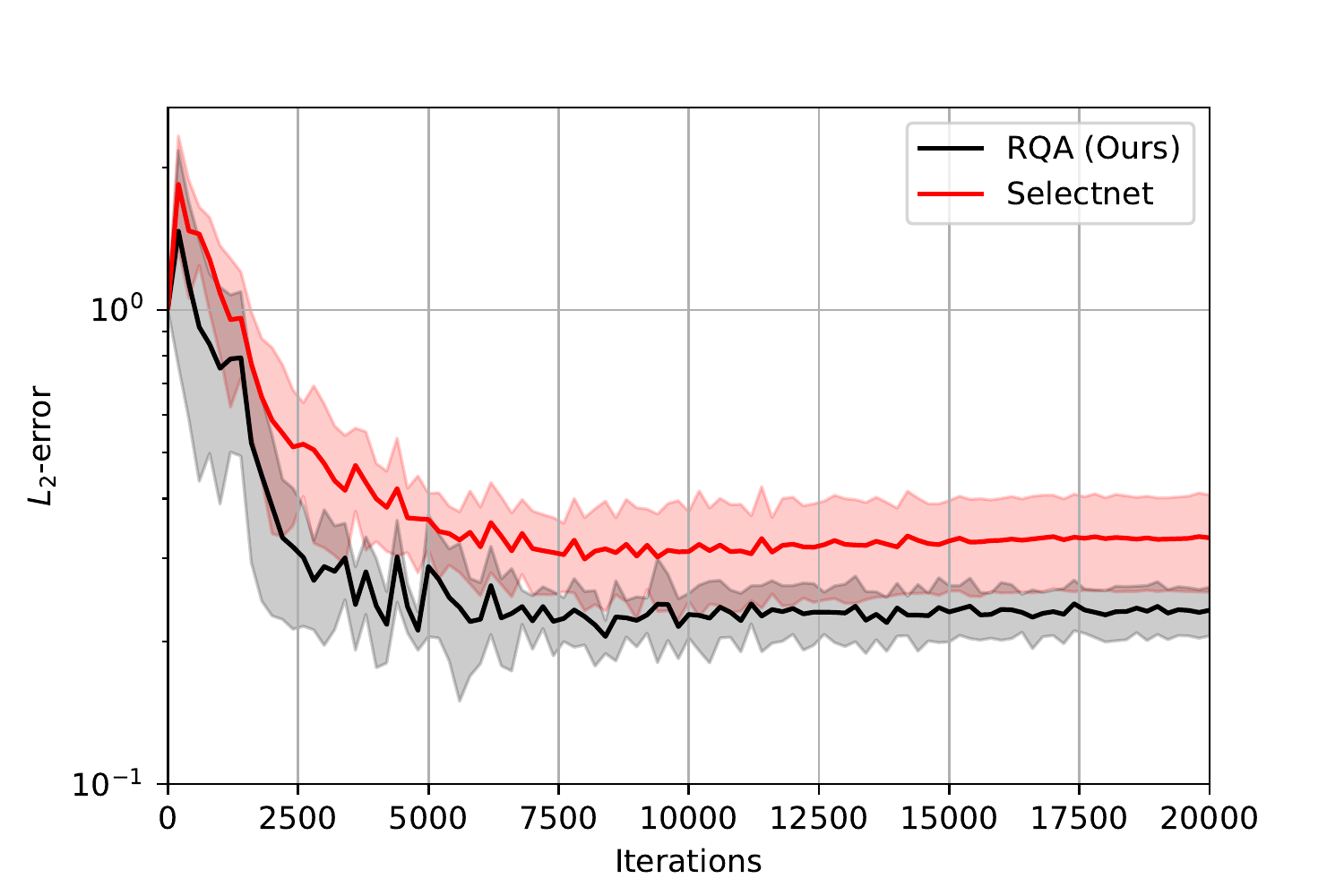}
    \caption{Comparison of $L_2$-error between RQA and Selectnet for 20 dim Allen-Cahn equation.}
    \label{Fig:3-20-l2}
    \end{minipage}
}
\subfigure{
\begin{minipage}[t]{0.4\linewidth}
\centering
    \includegraphics[width=\textwidth]{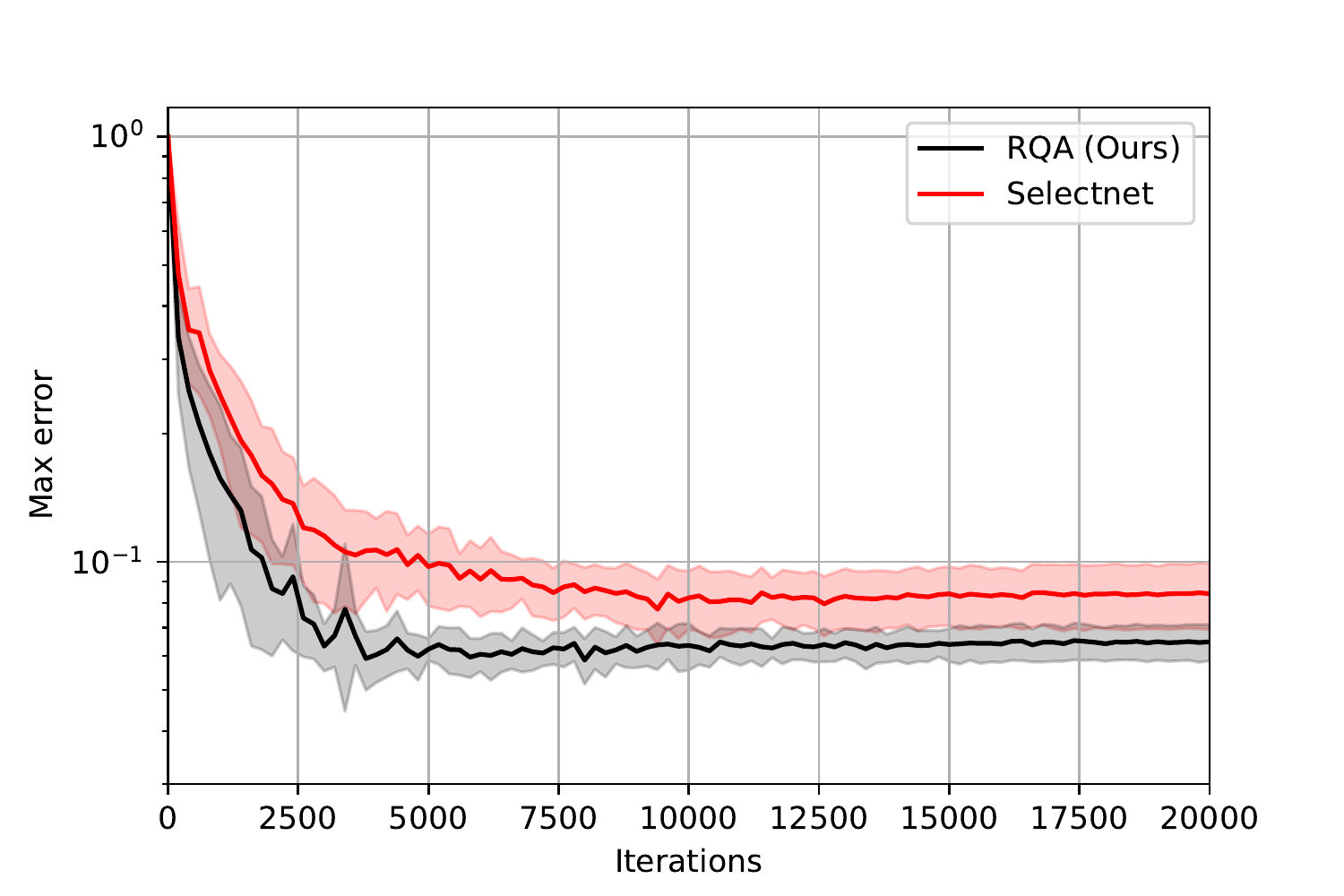}
    \caption{Comparison of absolute maximum error between RQA and Selectnet for 20 dim Allen-Cahn equation.}
    \label{Fig:3-20-m}
    \end{minipage}
}
\end{figure*}

\subsection{High-dimensional Problem}\label{sec:exp2}

In this section, we show how RQA performs for the high-dimensional case. Specifically, we compare RQA with Selectnet for the linear parabolic equation \eqref{eq:lpe} and Allen-Cahn equation \eqref{eq:ac} of 20 dimensions. 

The two methods are trained for $20,000$ iterations over 5 repeated experiments for both problems. In every single run, we set the initialization for the solution network to be the same. The training set and test set at each iteration are also the same. During the experiment, we record the $L_2$-error and absolute maximum error of each iteration.

Figure \ref{Fig:2-20-l2} and Figure \ref{Fig:2-20-m} show the comparison of the two methods on $L_2$-error and absolute maximum error for the linear parabolic equation. The solid line refers to the mean error and the shaded area is $[mean-std,mean+std]$ as above. The black line is our RQA and the red line is Selectnet. RQA has a smaller error than Selectnet in both figures.

For the 20-dim Allen-Cahn equation, the average $L_2$-error and absolute maximum error of the two methods are displayed in 
Figure \ref{Fig:3-20-l2} and Figure \ref{Fig:3-20-m} respectively. As before, our RQA has better performance than Selectnet for this problem.

% (Not finished yet)
% For this example, we compare the methods for 20-dim problem and 20-dim problem. The results for the 2-dim case are shown in Fig. \ref{Fig:1-2-l2} and Fig. \ref{Fig:1-2-m}. From the figure, we can see that the proposed has the best $l_2$ error. For theabsolute maximum loss, although the no quantile adjustment version has the best performance, the gap with our method is not large. Selectnet outperforms the basic method as it is supposed to be for this case.

% Results for the 10-dim case are shown in Fig. \ref{Fig:1-10-l2} and Fig. \ref{Fig:1-10-m}. Notice that our method has the lowest $l_2$ error and absolute maximum loss. Meanwhile, the no quatile adjustment version performs the worst. Also, for this case, Selectnet has almost the same performance as the basic model. Notice that the data is sufficient for the previous case but not sufficient for this case. Selectnet is more easily influenced by the datasize but our method is not.

% The result for this case are shown in Fig. \ref{Fig:2-10-l2} and Fig. \ref{Fig:2-10-m}. Our method perform the best for both $l_2$ loss and the absolute maximum loss. Also, the performance of no quantile version has the worse error, which emphasis the importance of avoiding singularity.

\subsection{Discussion of parameters}\label{sec:exp3}
It has been shown in the previous sections that the proposed RQA performs well in solving PDE adaptively. However, we have not discussed hyper-parameters. The two important parameters are the $p$ value and the quantile we used to adjust weights for data points with large residuals.

To study the influence of these two parameters, we consider the elliptic equation as follows.
\begin{equation}
\label{eq:prob1}
\begin{split}
\mathcal{N}_3[u(x)] &=f(x),  \text { in } \Omega:=\{x:\|x\|_2<1\}, \\
u &=g(x),  \text { on } \partial \Omega,
\end{split}
\end{equation}
with 
\begin{align}
\mathcal{N}_3[u(x)]=&-\nabla \cdot(a(x) \nabla u)+|\nabla u|^{2},\\
f(x)=&\frac{5\pi \|x\|_2}{4}\cos(I)\left(1-\|x\|_2\right)^{1.5}\notag\\
&-(1+0.5\|x\|_2^2)L(x)\\
&+\frac{25}{16}\pi^2 \cos(I)^2(1-\|x\|_2)^3,\notag\\
g(x)=&0
\end{align}
where
\begin{align}
I(x)=&\frac{\pi}{2}\left(1-\|x\|_2\right)^{2.5},\\
L(x)=&\frac{-5\pi (d-1)\cos (I(x))(1-\|x\|_2)^{1.5}}{4\|x\|_2}\notag\\&-\frac{25\pi^2}{16}\sin(I(x))\left(1-\|x\|_2\right)^3\\&+\frac{15\pi}{8}\cos(I(x))\left(1-\|x\|_2\right)^{0.5},\notag\\
a(x)=&1+\frac{1}{2}\|x\|_2^2,
\end{align}
here $d$ is the dimension of $x$. The exact solution is 
\begin{align}
u(x)=\sin(I(x)).
\end{align}

We use RQA with different settings to solve this problem with a dimension of 5. Specifically, the $p$ value is chosen to be $p=2.5,3,4,5,6$. Larger $p$ means more emphasis on the data points with larger residuals, however, more singularity may be introduced. So the selection of $p$ needs to be treated carefully.

For the weights larger than $90\%$ quantile, we also study how "significant" we need to adjust them. Setting them to a large value may not eliminate singularity while setting them to a small value may make the adaptivity invalid. In the experiment below, we consider adjusting the large weights to $10\%$, $25\%$, $50\%$, $75\%$, and $90\%$ quantile. 

The selection for $p$ and quantile are paired with one another. For each ($p$, quantile) pair, we conduct RQA on 5 different runs. In Table \ref{table:1}, we report the average final $L_2$-error for every problem setting. The best results of each column in emphasized with bold text.

\begin{table*}[htb]   
\begin{center}   
\caption{{\upshape Average $L_2$-error for different $p$ and different quantile in RQA.}}  
\label{table:1} 
\begin{tabular}{lccccc}    
\hline\hline   & $10\%$ & $25\%$ & $50\%$ &$75\%$ &$90\%$  \\ 
\hline\hline   $p=2.5$& $\boldsymbol{9.199\times10^{-2}}$ & $\boldsymbol{6.802\times10^{-2}}$ &$1.335\times10^{-1}$ &$5.430\times10^{-2}$& $3.276\times10^{-2}$ \\  
\hline   $p=3$ &$1.375\times10^{-1}$ &$1.590\times10^{-1}$&$7.503\times10^{-2}$ &$2.368\times10^{-2}$&$3.274\times10^{-2}$\\
\hline   $p=4$ &$1.496\times10^{-1}$&$1.375\times10^{-1}$&$\boldsymbol{9.449\times10^{-3}}$&$2.389\times10^{-2}$&$2.65\times10^{-2}$\\
\hline   $p=5$ &$3.524\times10^{-1}$&$1.702\times10^{-1}$&$1.962\times10^{-2}$&$\boldsymbol{1.061\times10^{-2}}$&$1.392\times10^{-2}$\\
\hline   $p=6$&$3.668\times10^{-1}$&$3.417\times10^{-1}$&$9.424\times10^{-2}$&$1.320\times10^{-2}$&$\boldsymbol{1.345\times10^{-2}}$\\
\hline\hline 
\end{tabular}   
\end{center}   
\end{table*}
From the table, we can observe an interesting result that the best error appears on the diagonal of the table. In other words, the best choice for $p$ and quantile depend on each other. A smaller $p$ value should be paired with a smaller quantile and a larger $p$ value should be paired with a larger quantile.

This observation can be explained by our previous discussion that the function of quantile adjustment is to balance between applying adaptivity and preventing singularity. For smaller $p$ values, where the weight is not large, we only need to compensate the adjusted weight with a small value. So the performance of $10\%$ quantile is already good enough. However, for larger $p$ values, in which the weights to be adjusted are already very large, a larger value is needed. So $90\%$ quantile works the best.

Also, from Table \ref{table:1}, we see that the best $L_2$-error is the $p=4$ and $50\%$ quantile pair, which suggests that to apply RQA, one may start from a moderate quantile and a $p$ value neither too large nor too small.

\begin{figure*}[htbp]
\centering
\subfigure{
\begin{minipage}[t]{0.3\linewidth}
\centering
    \includegraphics[width=\textwidth]{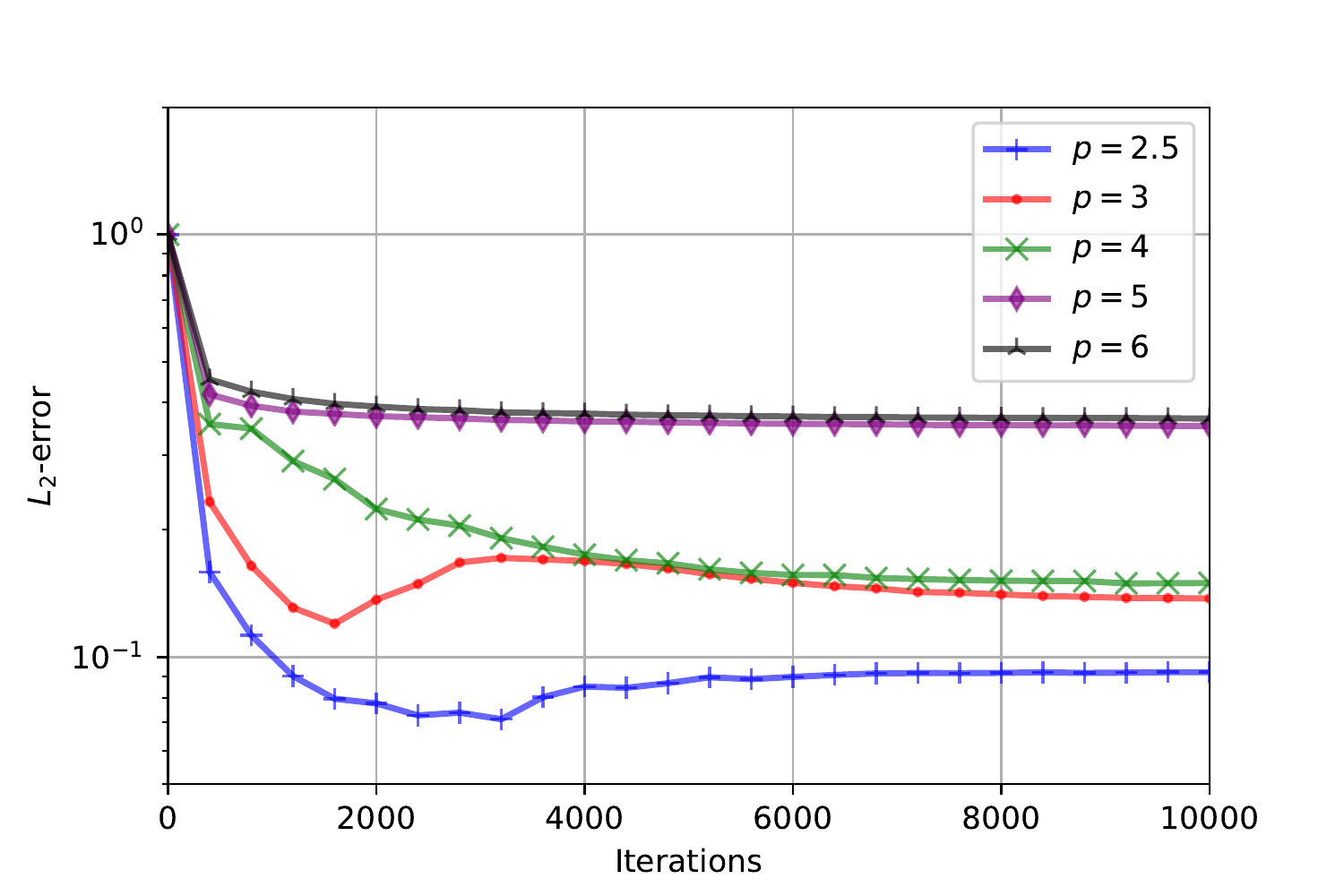}
    \caption{Quantile: $10\%$.}
    \label{fig:q10}
    \end{minipage}
}
\subfigure{
\begin{minipage}[t]{0.3\linewidth}
\centering
    \includegraphics[width=\textwidth]{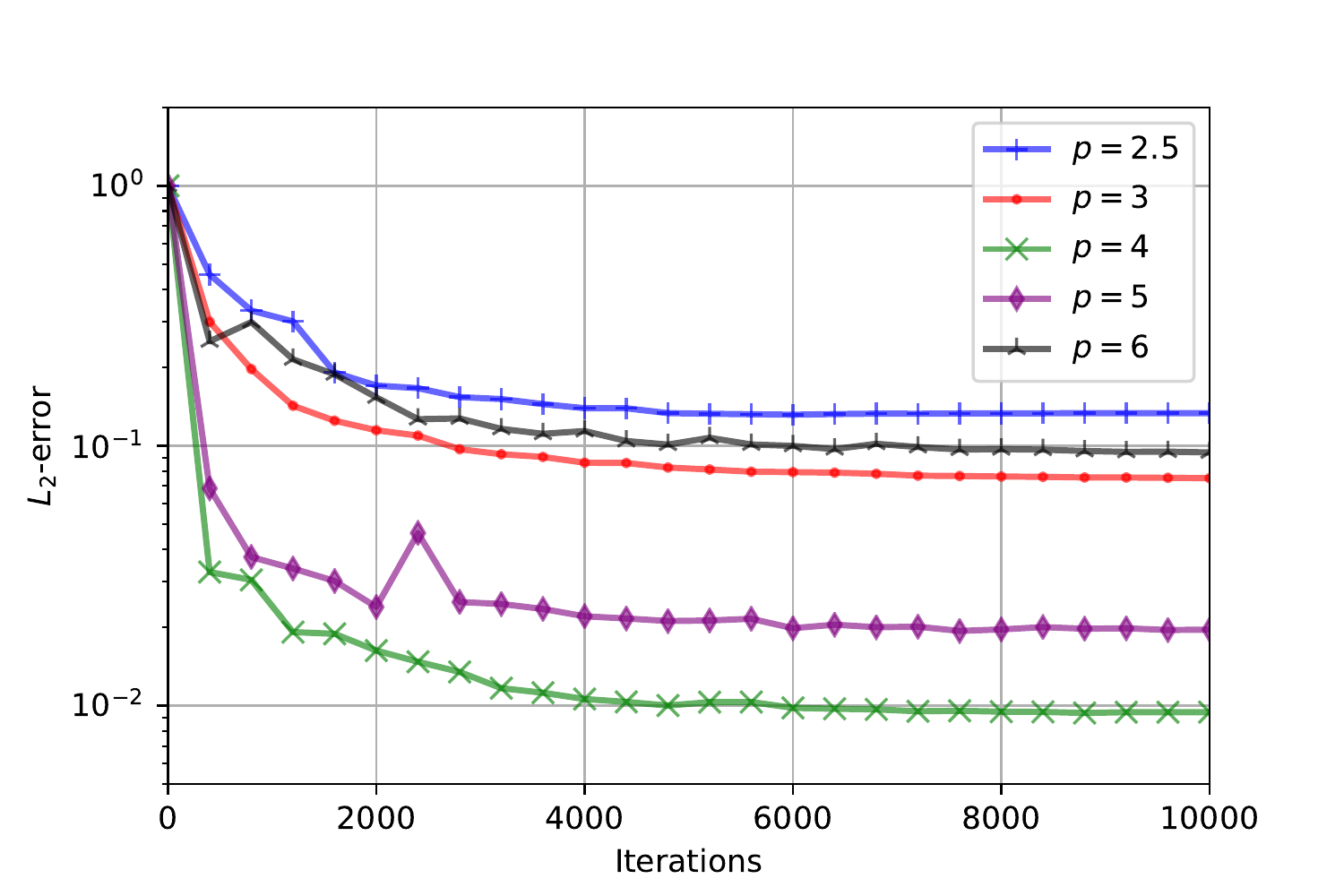}
    \caption{Quantile: $50\%$.}
    \label{fig:q50}
    \end{minipage}
}
\subfigure{
\begin{minipage}[t]{0.3\linewidth}
\centering
    \includegraphics[width=\textwidth]{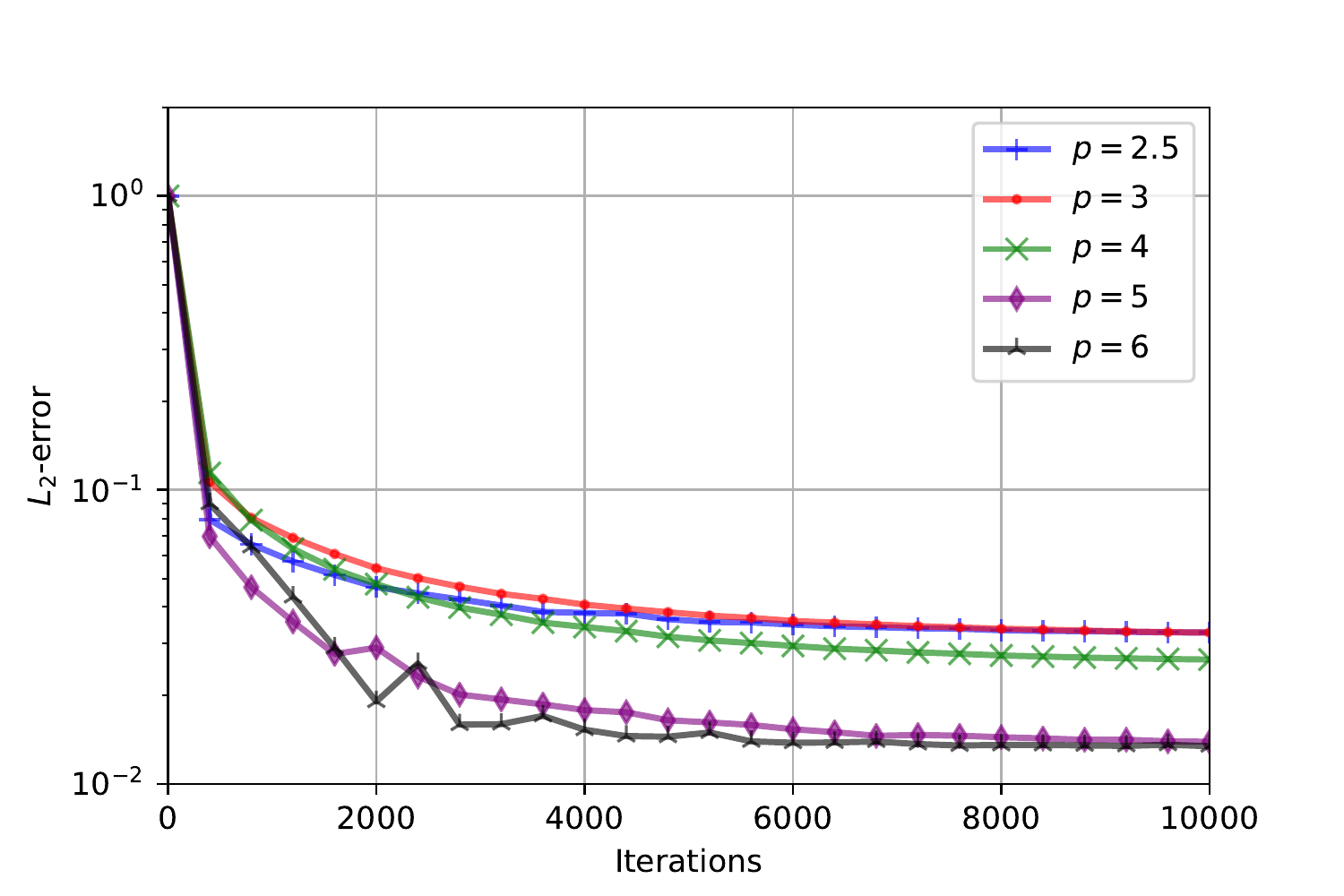}
    \caption{Quantile: $90\%$}
    \label{fig:q90}
    \end{minipage}
}
\end{figure*}

To offer a more intuitive impression of the selection of parameters, we plot the average $L_2$-error at each iteration for the first, third, and fifth column of Table \ref{table:1} in Figure \ref{fig:q10}, Figure \ref{fig:q50} and Figure \ref{fig:q90} respectively.  

The three figures demonstrate the training process for various $p$ values when quantile $10\%$, $50\%$, and $90\%$ are applied. For quantile set to $10\%$, $p=2.5$ works the best. When the $p$ value gets larger, its performance gets worse. In the quantile $50\%$ case, $p$ with moderate value, say $p=4$, converges the fastest. Neither small $p$ value, for example, $p=2.5$, nor large $p$ value, say $p=6$, has good performance. When the quantile gets to $90\%$, the performance of different $p$ values is completely different from quantile $10\%$. Large $p$ values ($p=5,6$) now have better performance and small $p$ values ($p=2.5,3$) do not work well.

Finally, we emphasize that the above conclusion can be generalized to other adaptive methods except for $L_p$-norm induced weight. The $p$ here actually reflects the ``intensity" of the adaptivity.

%%%%%%%%%%%%%%%%%%%%%%%%%%%%%

\section{Conclusion}
\label{sec:con}

% In this work, we design an auxiliary network free strategy for solving PDE adaptively. The adaptivity is conducted by residual-based weight and to avoid singularity, we introduced a quantile regression adjustment. This approach replace the largest weights by the median, thus guarantee that all data points can contribute to the loss function. Experiments results show that our method outperform Selectnet and the no qualtile version. Future work involves comparing our method with more PDE solvers, for example: WAN, and testing the method on more PDE examples.

In this work, we propose an iterative reweighting technique for the adaptive training of PINN. The weights are chosen based on the distribution of the residual with an effective adjustment step to mitigate the long tail. 
% We demonstrate that the performance of any adaptive method is
% highly related to the tail behavior of the residual during the training process. 
The proposed method, RQA,  
adjusts the weights by resetting those values above a certain quantile (for example the $90\%$ quantile) toward a median value to 
achieve the balance between very few data points with large residuals and the majority of data points with small residuals. Experimental results show that RQA has better performance and faster convergence than the baseline methods for a few benchmark PDEs.  Also, the selection of the quantile should be apropos to the selection of $p$ (the ``intensity" of adaptivity). Small $p$ is advised to cooperate with a small quantile threshold for RQA. We anticipate a further exploration for the generalization of RQA applied to other high-dimensional PDE solvers beyond PINN and studying how our method can deal with failure mode for PINN. 

\section*{Acknowledgement}

The authors thank the support of Hong Kong RGC GRF grants 11307319, 11308121, 11318522 and NSFC/RGC Joint Research Scheme (CityU 9054033).

% In this work, we propose that the maximizer for the selection network in Selectnet is not computationally efficient and thus not necessary, and can be substituted by the $L_p$-norm of the residual. Considering that the $L_p$-norm of the residual may vary severely, in other words, the distribution of weights resulting from the normalization of $L_p$-norm of the residual is case sensitive and may be too "singular" by focusing on several points whose residuals are in dominant states and may ignore the demanding of gradient descend of majority points in the very beginning of training process, that is when the "long tail" phenomenon appears. Then the loss function will pay more attention to the "important" data points and neglect the "unimportant" data points. To fix this, Residual-Quantile Adjustment algorithm (RQA) is proposed to balance the weights between "important" and "unimportant" data points. For our experiments, RQA has better performance and faster convergence than the baseline methods on benchmark PDEs. Also, the selection of the quantile should be adapted to the selection of the $p$. Small $p$ is advised to cooperate with small quantile for RQA. We anticipate further exploration for the generalization of RQA applied in the other residual-based methods. Theoretical guidance for the selection of $p$ and quantile is still highly disputable in the future. 

% \section*{Acknowledgment}
% Xiang Zhou acknowledges the support of Hong Kong GRF 9043414. Zhiqiang Cai and Jiayue Han acknowledges the support of UGC for PhD candidates.
\bibliography{ref}
\bibliographystyle{ieeetr} %alpha, siam

% \bibitem{b1} G. Eason, B. Noble, and I. N. Sneddon, ``On certain integrals of Lipschitz-Hankel type involving products of Bessel functions,'' Phil. Trans. Roy. Soc. London, vol. A247, pp. 529--551, April 1955.
% \bibitem{b2} J. Clerk Maxwell, A Treatise on Electricity and Magnetism, 3rd ed., vol. 2. Oxford: Clarendon, 1892, pp.68--73.
% \bibitem{b3} I. S. Jacobs and C. P. Bean, ``Fine particles, thin films and exchange anisotropy,'' in Magnetism, vol. III, G. T. Rado and H. Suhl, Eds. New York: Academic, 1963, pp. 271--350.
% \bibitem{b4} K. Elissa, ``Title of paper if known,'' unpublished.
% \bibitem{b5} R. Nicole, ``Title of paper with only first word capitalized,'' J. Name Stand. Abbrev., in press.
% \bibitem{b6} Y. Yorozu, M. Hirano, K. Oka, and Y. Tagawa, ``Electron spectroscopy studies on magneto-optical media and plastic substrate interface,'' IEEE Transl. J. Magn. Japan, vol. 2, pp. 740--741, August 1987 [Digests 9th Annual Conf. Magnetics Japan, p. 301, 1982].
% \bibitem{b7} M. Young, The Technical Writer's Handbook. Mill Valley, CA: University Science, 1989.
% \end{thebibliography}
% \vspace{12pt}
% \color{red}
% IEEE conference templates contain guidance text for composing and formatting conference papers. Please ensure that all template text is removed from your conference paper prior to submission to the conference. Failure to remove the template text from your paper may result in your paper not being published.

\end{document}